\documentclass[journal, compsoc]{IEEEtran}
\usepackage{cite}
\usepackage{amsmath,amssymb,amsfonts}
\usepackage{algorithmic}
\usepackage{array}
\usepackage{graphicx}
\usepackage{textcomp}
\usepackage[table]{xcolor}
\usepackage{xcolor}
\usepackage{orcidlink}
\usepackage{multicol}
\usepackage{multirow}

\usepackage[inline]{enumitem}
\usepackage{booktabs}
\usepackage{pifont}
\usepackage{arydshln}

\def\BibTeX{{\rm B\kern-.05em{\sc i\kern-.025em b}\kern-.08em
    T\kern-.1667em\lower.7ex\hbox{E}\kern-.125emX}}
\usepackage{listings}

\definecolor{codegreen}{rgb}{0,0.6,0}
\definecolor{codegray}{rgb}{0.5,0.5,0.5}
\definecolor{codepurple}{rgb}{0.58,0,0.82}
\definecolor{backcolour}{rgb}{0.95,0.95,0.92}

\lstdefinestyle{mystyle}{
    backgroundcolor=\color{backcolour},   
    commentstyle=\color{codegreen},
    keywordstyle = {\color{magenta}},
    keywordstyle = [2]{\color{lime}},
    keywordstyle = [3]{\color{yellow}},
    keywordstyle = [4]{\color{teal}},
    numberstyle=\tiny\color{codegray},
    stringstyle=\color{codepurple},
    basicstyle=\ttfamily\footnotesize,
    breakatwhitespace=false,         
    breaklines=true,                 
    captionpos=b,                    
    keepspaces=true,                 
    numbers=left,                    
    numbersep=5pt,                  
    showspaces=false,                
    showstringspaces=false,
    showtabs=false,                  
    tabsize=2
}
\newcommand{\cmark}{\ding{51}}  % check mark
\newcommand{\xmark}{\ding{55}}  % cross mark

\begin{document}

\title{CoDAT: Collaborative Dual-Attention Transformer with Low-Cost Temporal Modeling for Efficient Edge Action Recognition  \\
}
\author{Novendra Setyawan, \IEEEmembership{Student, IEEE}, Chi-Chia Sun, \IEEEmembership{Member, IEEE}, Mao-Hsiu Hsu, \IEEEmembership{Member, IEEE} \\
Wen-Kai Kuo, \IEEEmembership{Member, IEEE},
Jing-Ming Guo, \IEEEmembership{Fellow, IEEE},
Jun-Wei Hsieh, \IEEEmembership{Senior Member, IEEE}\\
\IEEEauthorblockA{} 
\thanks{
This research was supported by the National Science and Technology Council, Taiwan through Grant Number NSTC-113-2221-E-305-018-MY3. \textit{(Corresponding Author: Chi-Chia Sun)}

Novendra Setyawan is with Department of Electro-Optics, National Formosa University, Huwei 632301, Taiwan and also with Department of Electrical Engineering University of Muhammadiyah Malang, Malang 65144, Indonesia. ({e-mail: novendra@umm.ac.id})

Chi-Chia Sun is with Department of Electronic and Computer Engineering, National Taiwan University of Science and Technology, Taipei 106335, Taiwan ({e-mail: chichiasun@mail.ntust.edu.tw}). 

Mao-Hsiu Hsu and Wen-Kai Kuo are with Department of Electro-Optics, National Formosa University, Huwei 632301, Taiwan.

Jing-Ming Guo is with Department of Electrical Engineering, National Taiwan University of Science and Technology, Taipei 106335, Taiwan. 

Jun-Wei Hsieh is with College of Artificial Intelligence and Green Energy, National Yang Ming Chiao Tung University, Tainan 711010, Taiwan.
}}

\maketitle
\markboth{Accepted on IEEE Internet of Things Journal}
{Setyawan \etal{}: CoDAT}

\begin{abstract}
Real-time human action recognition on Internet-of-Things (IoT) edge devices requires models that capture rich spatio-temporal cues within strict latency, memory, and power envelopes. Current 3D CNNs, video transformers, and shift-based ViT deliver high accuracy but come at computational costs that preclude edge IoT deployment. This paper proposes \emph{CoDAT}, a Collaborative Dual-Attention Transformer that replaces conventional multi-head attention with a lightweight dual-branch module: Spatial Convolutional Attention (SCA) for local aggregation and Strided Single-Head Attention (SSHA) for global context. SSHA jointly compresses the spatial resolution and channel dimensions of the query, key, and value tensors via stride-based sparse projection, then fuses the resulting global and local features at a markedly reduced cost. To enable temporal communication across frames, a parameter-free TShift module is embedded in each block. Extensive experiments on Jetson AGX Orin and Raspberry Pi 5 demonstrate that CoDAT achieves an energy-accuracy balance in both image and action recognition. On ImageNet-1K, CoDAT-M runs $2\times$ faster than EfficientViT$_{384}$ and FastViT-S12 at comparable accuracy, and CoDAT-L matches ViT-S with $3\times$ fewer parameters at $2\times$ higher throughput. On Kinetics-400 and MA-52, CoDAT achieves competitive Top-1 accuracy against state-of-the-art CNN, transformer, and hybrid baselines while running up to $2.9\times$ faster than VSwin-T, $2\times$ faster than ViT-Temporal-Shift variants, and $5\times$ faster than UniFormer-B. On UCF-101, CoDAT-S$_{384}$ matches TokShift and LAPS while being  $6\times$ faster and requiring up to $13\times$ fewer FLOPs, establishing an efficiency-accuracy balance for real-time action recognition in edge IoT perception systems. Code is available at \href{https://github.com/novendrastywn/CoDAT}{\textit{https://github.com/novendrastywn/CoDAT}}.
\end{abstract}
\section{Introduction}
\label{sec:intro}
\begin{figure}
    \centering
    \includegraphics[width=\columnwidth]{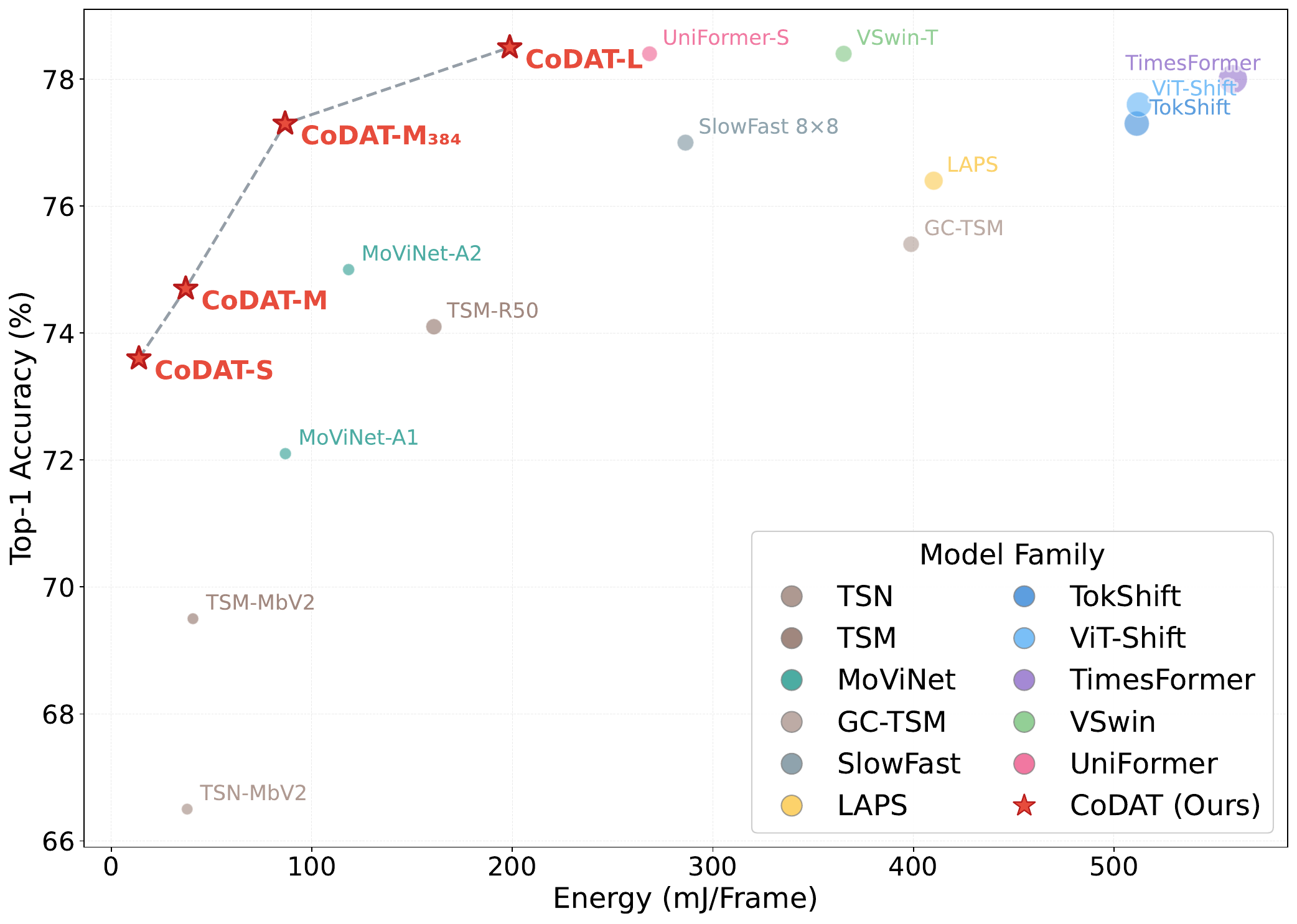}
    \caption{Comparison of our proposed CoDAT model with SOTA methods.  Accuracy v.s. Energy per Frame. The upper-left of the plot represents higher accuracy is achieved with minimal energy.}
    \label{fig:acc_en}
\end{figure} 
\IEEEPARstart{H}{uman} action recognition (HAR) has emerged as a central problem in IoT perception systems, with wide-ranging applications such as video surveillance~\cite{sun2019vu, khan2025strack}, sports analytics \cite{muhammad2021ai}, healthcare monitoring \cite{wang2023contactless}, human-computer interaction \cite{li2025optimization}, and autonomous driving \cite{11134493}. The goal of action recognition is to automatically identify human activities from video or multi-frame sequences by understanding both spatial appearance and temporal motion cues. The significance of this task lies in its potential to endow machines with a deeper perceptual understanding of dynamic environments~\cite{khan2026tracenet,baz2025hamot}, enabling intelligent decision-making in real-world contexts. However, action recognition remains challenging due to the complex variability of motion patterns, viewpoint changes, occlusions, and environmental clutter. These factors demand models capable of efficiently integrating spatio-temporal information while maintaining real-time inference capability, which is an increasingly critical requirement for modern vision systems deployed on energy-constrained IoT edge and embedded devices.

Over the past decade, convolutional neural networks (CNNs) have significantly advanced visual recognition. Since the introduction of AlexNet \cite{krizhevsky2012imagenet}, CNNs have achieved major breakthroughs in image classification \cite{setyawan2025facelivt, setyawan2026facelivtv2}, object detection \cite{10208307, wang2024smiletrack, khan2026lighttrack}, and semantic segmentation \cite{11299097}. Their inherent inductive biases, including local connectivity, translation equivariance, and hierarchical representation learning, enable CNNs to extract discriminative spatial features with high computational efficiency. These strengths naturally extend into action recognition, where 2D CNNs, two-stream networks, and optical-flow-based methods provide strong baselines for capturing appearance and short-term motion information. Subsequently, the architectures of 3D CNNs further accelerated progress, leading to significant improvements in accuracy and robustness across video benchmarks \cite{soomro2012ucf101, kay2017kinetics, guo2024benchmarking}, firmly establishing CNNs as the dominant paradigm in early video understanding research.

Despite their success, 3D CNN-based action recognition models suffer from inherent limitations. Architectures such as I3D~\cite{carreira2017quo} and SlowFast~\cite{feichtenhofer2019slowfast} model spatio-temporal features by extending convolution operations into the temporal dimension or by factorizing 3D kernels into spatial and temporal components. However, these models typically require heavy computation, large memory footprints, and prolonged training due to dense 3D operations. Although decomposed variants such as STANet~\cite{li2023spatio} and AGPN~\cite{chen2023agpn} reduce complexity by separating spatial and temporal convolutions, they remain constrained by their local receptive fields and struggle to capture long-range temporal dependencies, which are essential for recognizing complex or subtle human actions. More critically, the high computational and energy costs of 3D CNNs make them impractical for deployment under the stringent power, memory, and latency budgets of IoT edge devices.

Meanwhile, Vision Transformers (ViTs) \cite{dosovitskiy2020image}, originally developed for image recognition, have demonstrated strong global reasoning capabilities through self-attention, motivating their adoption in video understanding. Transformer-based models such as TimeSformer~\cite{bertasius2021space}, VTN~\cite{neimark2021video}, Video Swin Transformer~\cite{liu2022video}, and UniFormer~\cite{li2023uniformer} apply spatio-temporal attention or extend the transformer architecture from 2D to 3D, achieving strong performance on large-scale datasets. However, the quadratic complexity of multi-head self-attention incurs substantial computational and memory overhead, limiting their suitability for resource-constrained environments. To mitigate this cost, shift-based variants such as TokShift~\cite{zhang2021tokenshift}, LAPS~\cite{zhang2022long}, and ViTs-Shift~\cite{zhang2024temporal} introduce temporal channel shifting into ViTs to simulate inter-frame interaction without explicit temporal attention, enabling lightweight temporal modeling at negligible additional cost. Nevertheless, even with shift-based temporal processing, the underlying ViT backbones remain computationally demanding and energy-intensive, rendering them impractical for real-time inference on low-power mobile and edge platforms.

To address these limitations, we propose CoDAT (\textbf{Co}llaborative \textbf{D}ual-\textbf{A}ttention \textbf{T}ransformer), an energy-efficient backbone that jointly captures global and local spatial representations through a unified attention mechanism. The core of CoDAT is the Collaborative Dual-Attention (CoDA) module, which operates two lightweight branches in parallel: Strided Single-Head Attention (SSHA) for efficient global spatial dependency modeling and Spatial Convolutional Attention (SCA) for local spatial aggregation. Rather than relying on simple concatenation, the two branches interact through a learnable cross-branch projection that enables complementary feature fusion. By employing single-head attention with strided spatial reduction alongside convolutional saliency cues, the CoDA module substantially reduces both memory access and arithmetic complexity, directly targeting the two major bottlenecks for real-time inference on resource-constrained IoT edge devices.
\begin{table*}[!ht]
\centering
\caption{Key architectural properties of representative human action recognition models.}
\label{tab:arch_comparison}
\setlength{\tabcolsep}{2.5pt}
\renewcommand{\arraystretch}{1.1} % Taller rows (default 1)
\begin{tabular}{m{1.9cm} ccccp{11.5cm}}
\hline
\multirow{2}{*}{\textbf{Methods}} & \multicolumn{2}{c}{\textbf{Context}} & \textbf{Low-Cost} & \textbf{Edge} & \multicolumn{1}{c}{\multirow{2}{*}{\textbf{Approach}}} \\
 & \textbf{Global} & \textbf{Local} & \textbf{Temporal} & \textbf{Suitable} & \\
\hline
TSM~\cite{lin2022tsm}
 & \xmark & \cmark &  \cmark  & \cmark
 & ResNet-based~\cite{he2016deep} 2D CNN with low-cost temporal channel shifting; lacks global attention. \\
SlowFast~\cite{feichtenhofer2019slowfast}
 & \xmark & \cmark &  \xmark  & \xmark
 & Dual-pathway 3D ResNet operating at slow and fast frame rates for multi-scale temporal reasoning. \\
TimeSformer~\cite{bertasius2021space}
 & \cmark & \xmark &  \xmark & \xmark
 & Pure ViT~\cite{dosovitskiy2020image} with divided space-time attention; high computational cost from full self-attention. \\
TokShift~\cite{zhang2021tokenshift}
 & \cmark & \xmark & \cmark & \xmark
 & ViT-based with low-cost temporal modeling; heavy backbone limits edge IoT deployment. \\
Video Swin~\cite{liu2022video}
 & \cmark & \cmark & \xmark & \xmark
 & Shifted-window 3D Transformer~\cite{liu2021swin} with local-to-global attention; heavy 3D window computation. \\
MoViNet~\cite{kondratyuk2021movinets}
 & \xmark & \cmark & \xmark & \cmark
 & Mobile 3D CNN with causal convolutions and stream buffers; designed for on-device inference. \\
UniFormer~\cite{li2023uniformer}
 & \cmark & \cmark & \xmark & \xmark
 & Unified 3D CNN in shallow and self-attention in deep layers for joint local--global modeling. \\
\hline
\textbf{CoDAT}
 & \cmark & \cmark & \cmark & \cmark
 & Collaborative Strided Single-Head Attention and Spatial Convolutional Attention with low-computation and parameter-free temporal modeling for efficient edge action recognition. \\
\hline
\end{tabular}
\end{table*}
Preserving strict efficiency constraints, we embed a zero-parameter temporal shift operation ($\mathrm{TShift}$) into the CoDAT architecture, enabling low-cost inter-frame communication that unifies spatial and temporal modeling without expensive attention along the temporal dimension, as summarized in Table~\ref{tab:arch_comparison}. As a result, CoDAT significantly alleviates the hardware burden associated with video understanding and is well-suited for deployment on low-power, latency-sensitive edge devices where computational budgets and energy constraints render conventional 3D CNNs and Transformers impractical.

The main contributions of this work are arranged and summarized as follows:
\begin{enumerate}
    \item We design Strided Single-Head Attention (SSHA), an efficient global spatial attention mechanism that applies strided spatial reduction in single-head self-attention, substantially lowering computational demand while preserving long-range dependency modeling.
    \item We propose the Collaborative Dual-Attention (CoDA) module, a joint global-local reasoning that combines SSHA with Spatial Convolutional Attention (SCA) to construct the CoDAT backbone network.
    \item We integrate a zero-parameter temporal shift mechanism ($\mathrm{TShift}$) into the CoDAT backbone to capture inter-frame dynamics with low-cost computation, yielding an end-to-end model for energy-efficient human action recognition.
    \item We validate CoDAT on standard image and video benchmarks, demonstrating a favorable accuracy-efficiency trade-off for real-time action recognition on resource-constrained IoT edge devices, as depicted in Fig.~\ref{fig:acc_en}.
\end{enumerate}

This paper is organized as follows. Section II reviews related work on CNNs for video understanding, transformer-based video recognition, and lightweight backbone approaches. Section III presents the proposed CoDAT architectures in detail. Section IV describes the experimental setup, benchmarks, evaluation results, and ablation studies. Finally, Section V concludes the paper and outlines future research directions.

%-------------------------------------------------------------------------

\section{Related Works}
\subsection{Human Action Recognition modalities}
HAR spans three primary sensing modalities: wearable sensor-based, WiFi signal-based, and video-based. Wearable sensor-based systems capture body motion via on-body sensors such as Inertial Measurement Unit (IMU)~\cite{fu2026deepsensemoe}, offering privacy preservation and lighting invariance, but require subject compliance and lack spatial context for fine-grained or multi-person recognition. While WiFi-based systems exploit CSI perturbations for device-free, through-wall recognition~\cite{liu2026sensor}, multi-modal fusion further improves robustness~\cite{wang2025deep}. However, CSI is restricted to coarse activity categories and degrades with environmental changes. Video-based systems provide rich spatio-temporal cues that support simultaneous multi-subject monitoring without on-body hardware and cover the fine-grained actions required for surveillance, healthcare, and sports scenarios. 
\begin{figure}[t!]
    \centering
    \includegraphics[width=\linewidth]{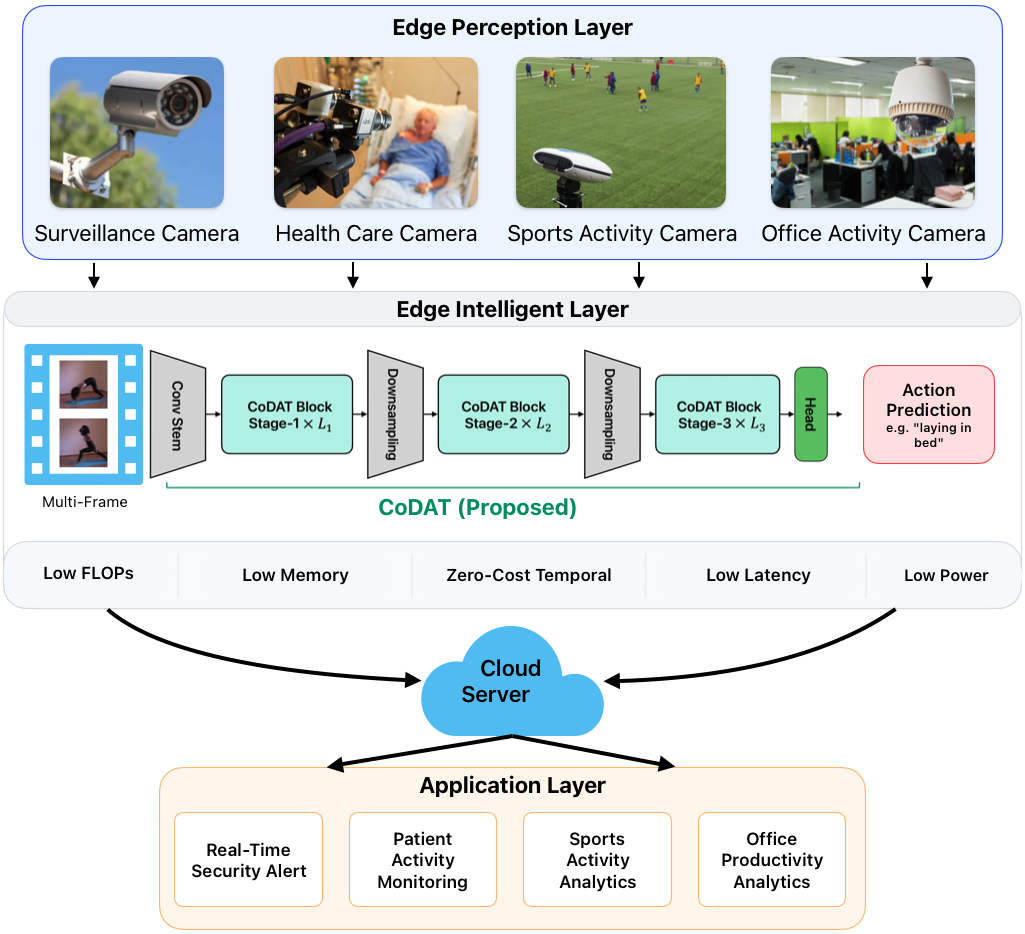}
    \caption{Cloud edge collaborative intelligent activity monitoring system with \textbf{CoDAT Architecture}}
    \label{fig:edge-ar}
\end{figure}
\begin{figure*}[t!]
    \centering
    \includegraphics[width=15cm]{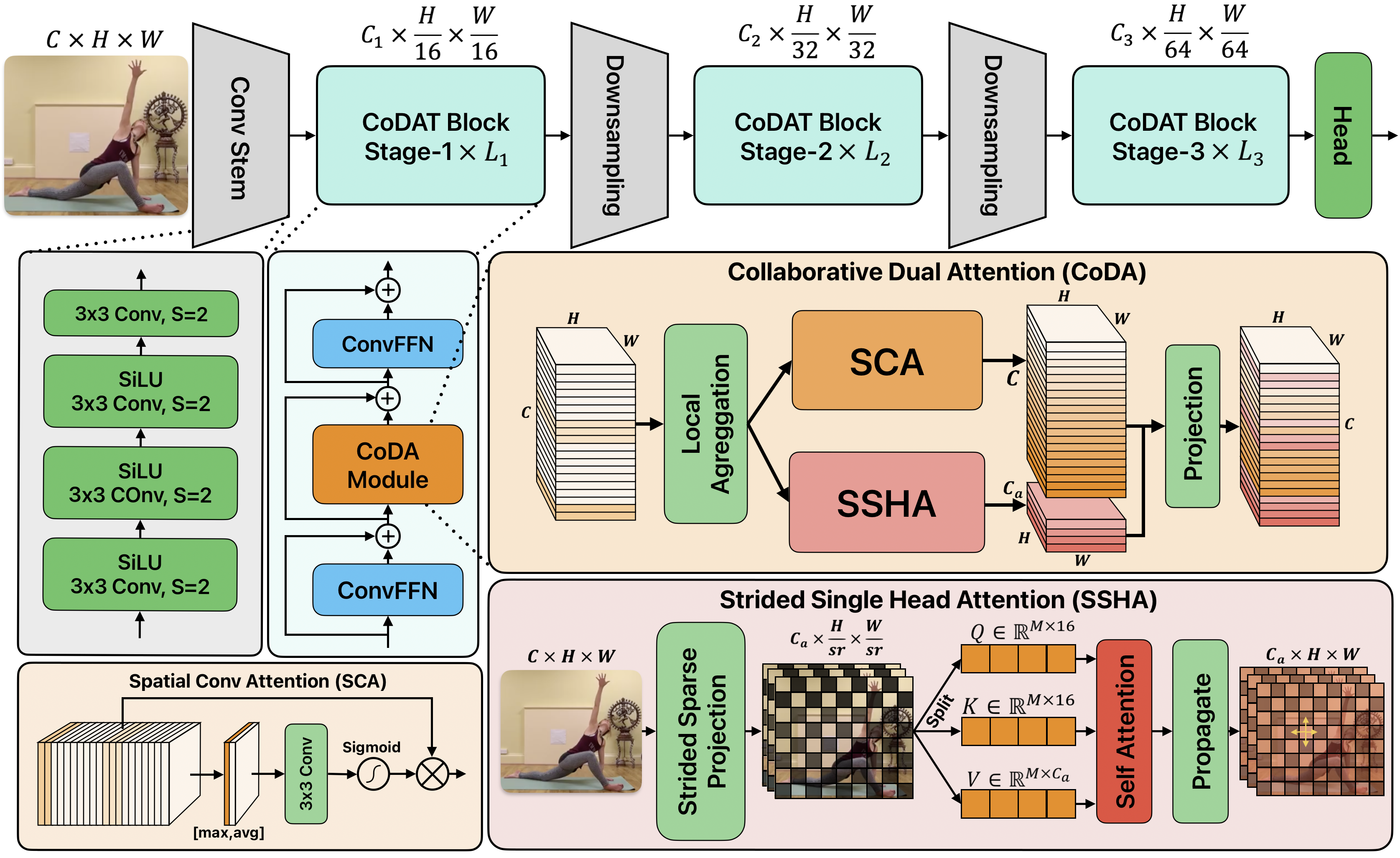}
    \caption{\textbf{The overall CoDAT Architecture}, CoDAT used pyramid architecture with 3 stages where each stage consist CoDA module, which operates two lightweight branches: Strided Single-Head Attention (SSHA) for efficient global spatial dependency modeling and Spatial Convolutional Attention (SCA) for local spatial aggregation.}
    \label{fig:codat_arch}
\end{figure*}
\subsection{CNN-Based Action Recognition}
CNNs have historically dominated video understanding owing to strong spatial representations and inductive biases such as locality and translation equivariance. 3D CNNs, including C3D~\cite{tran2015learning} and I3D~\cite{carreira2017quo}, jointly model spatial and temporal
information through spatio-temporal convolutions, achieving strong performance on benchmarks such as Kinetics-400~\cite{kay2017kinetics}. R(2+1)D~\cite{tran2018closer} decomposed 3D kernels into separate spatial and temporal operations to reduce computation, and SlowFast~\cite{feichtenhofer2019slowfast} refined temporal modeling through a dual-pathway design operating at different frame rates. Despite their effectiveness, 3D CNNs remain computationally heavy and memory-intensive, making them unsuitable for edge deployment.

A notable lightweight alternative is the Temporal Shift Module (TSM)~\cite{lin2019tsm}, which shifts a portion of feature channels along the temporal dimension to enable inter-frame interaction at low-cost and zero-parameters, achieving competitive accuracy with real-time performance on Jetson-class devices~\cite{lin2022tsm}. MoViNet~\cite{kondratyuk2021movinets} combines neural architecture search with causal convolutions for mobile-friendly 3D CNNs, while X3D~\cite{feichtenhofer2020x3d} progressively expands a small base architecture along multiple axes to identify efficient configurations.For the compressed-video domain, MTRFN~\cite{he2022mtrfn} introduces a multiscale temporal receptive field network that operates directly on MPEG motion vectors and residuals at edge servers, avoiding the cost of full decoding while capturing multi-granularity temporal patterns. Although these approaches reduce overhead, they still rely on learned temporal convolutions or multi-stream fusion, limiting their practicality under the strict computational and energy budgets of embedded IoT environments.

\subsection{Transformer-Based Action Recognition}
Transformers have reshaped video understanding by introducing global self-attention for long-range temporal modeling. Early video transformers extended ViTs through spatio-temporal factorization: VTN~\cite{neimark2021video} applies per-frame encoding with temporal aggregation; TimeSformer~\cite{bertasius2021space} decomposes attention into spatial-only and temporal-only components; Video Swin Transformer~\cite{liu2022video} extends shifted-window attention into the temporal dimension; and UniFormer~\cite{li2023uniformer} unifies local
convolution with global self-attention. TTSN~\cite{zhang2023temporal} combines temporal transformers with self-supervised auxiliary tasks to learn action representations. Despite these advances, explicit temporal attention remains computationally demanding.

Shift-based temporal transformers mitigate this overhead. TokShift~\cite{zhang2021tokenshift} shifts subsets of the CLS token across time steps to enable low-cost temporal reasoning, and ViT-Shift~\cite{zhang2024temporal} adapts TSM-style partial channel shifts to transformers while preserving the original ViT architecture. LAPS~\cite{zhang2022long} combines long-range dilated attention with periodic intra-head channel shifts. However, these methods still rely on standard ViT backbones with quadratic $\mathcal{O}(N^2)$ attention complexity, limiting their deployability on low-power edge devices.

\subsection{Lightweight Vision Transformers with Efficient Attention}
Efficient transformers have become critical for image and video understanding under resource constraints. Swin Transformer~\cite{liu2021swin} restricts attention to local shifted
windows, and PVT~\cite{wang2021pyramid, wang2022pvt} introduces spatial-reduction attention within a hierarchical pyramid. EdgeNeXt~\cite{maaz2022edgenext} computes attention along the channel dimension via transposed self-attention, while FastViT~\cite{vasu2023fastvit} leverages structural reparameterization for fast inference. EfficientViT~\cite{liu2023efficientvit} reduces memory movement through cascaded group attention, and EfficientFormer~\cite{li2023rethinking} re-evaluates attention design for
mobile deployment.

More recently, SHViT~\cite{yun2024shvit} showed that multi-head self-attention contains redundant patterns, motivating single-head attention with channel reallocation to achieve competitive accuracy at a reduced cost. S2AFormer~\cite{XuS2AFormer2025} proposes strip self-attention that jointly compresses spatial and channel dimensions. While these works individually address specific inefficiencies, they treat long token sequences, channel redundancy, and global--local representation as separate concerns, leaving a persistent accuracy--efficiency gap. Moreover, none are designed or evaluated for temporal video understanding, leaving their applicability to action recognition on edge devices unexplored.

\begin{figure*}[!t]
    \centering
    \includegraphics[width=\linewidth]{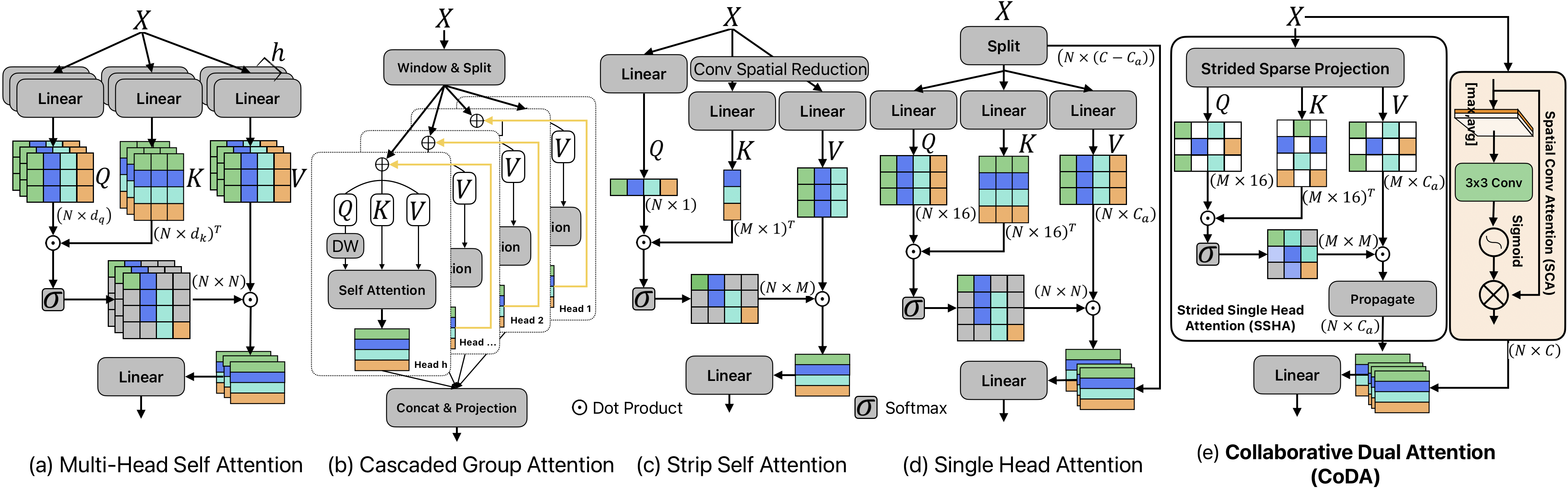}
    \caption{Comparison of different self-attention mechanisms: (a)~Multi-Head Self-Attention (MHSA)~\cite{touvron2021training}, (b)~Cascaded Group Attention~\cite{liu2023efficientvit}, (c)~Strip Self-Attention~\cite{XuS2AFormer2025}, (d)~Single-Head Attention~\cite{yun2024shvit}, (e)~Our proposed CoDA, which combines Strided Single-Head Attention that jointly compresses spatial and channel dimensions with lightweight Spatial Convolutional Attention to achieve an efficient design while preserving global dependencies.}
    \label{fig:coda_ssha_bench}
\end{figure*}

\subsection{Motivation}
Across the reviewed approaches, a persistent gap remains between temporal modeling capability and IoT hardware efficiency. In spite of recent IoT-oriented work such as TLEE~\cite{wang2023tlee}, which addresses this through temporal-wise and layer-wise early exiting, it sacrifices peak accuracy for efficiency. Shift-based temporal methods~\cite{lin2019tsm, zhang2021tokenshift, zhang2024temporal, zhang2022long} provide low-cost temporal reasoning but have been applied exclusively to heavy backbones such as ResNet and ViT, which are unsuitable for edge deployment. Conversely, recent lightweight backbone designs~\cite{liu2023efficientvit, vasu2023fastvit, yun2024shvit, li2023rethinking, XuS2AFormer2025} demonstrate that efficient spatial modeling is achievable, but none address temporal reasoning or have been validated for video action recognition under IoT constraints. The proposed CoDAT bridges this gap by combining Strided Single-Head Attention and Spatial Convolutional Attention in a unified Collaborative Dual-Attention module, augmented by low-cost temporal module for energy-efficient video understanding on resource-constrained edge devices. Fig.~\ref{fig:edge-ar} illustrates the deployment pipeline, where edge cameras stream multi-frame sequences directly to CoDAT on a low-power accelerator for real-time on-device action prediction, without offloading inference to a cloud server.
\section{Proposed Method}
As shown in Fig.~\ref{fig:codat_arch}, the core idea behind CoDAT is to combine an efficient yet expressive transformer block, CoDA, with low-cost temporal modeling for energy-efficient video understanding on edge devices. 

\subsection{Collaborative Dual-Attention Module (CoDA)} 
\label{sec:coda}
The CoDA module decomposes spatial feature interaction into local convolutional aggregation and global attention reasoning to efficiently capture both local spatial cues and global contextual relationships within a unified framework. CoDA reduces computational costs both spatially and channel-wise while preserving global--local feature representation through two components: \textit{Strided Single-Head Attention (SSHA)}, which compresses spatial token sequences via strided projection for efficient global dependency modeling, and \textit{Spatial Convolutional Attention (SCA)}, which provides local spatial aggregation in a parallel dual-branch design. This module addresses two key inefficiencies in existing transformer blocks: (1) the redundancy of multi-head attention and (2) the limited local sensitivity of pure self-attention. By executing these two operations in parallel and merging their outputs through a learnable projection, CoDA enables complementary feature propagation while maintaining computational efficiency suitable for real-time inference on energy-constrained edge devices.

Suppose the input feature map $X \in \mathbb{R}^{C \times H \times W}$ with number of tokens $N=H\times W$ and $C$ channel dimension is first processed by a lightweight Local Aggregation step following~\cite{chu2021conditional, vasu2023fastvit, liu2023efficientvit, yun2024shvit}, which enhances spatial consistency through dynamically generated kernels conditioned on the local neighborhood, preparing the features for parallel processing by SCA and SSHA.
The Local Aggregation computes the input features as described in Eq.~(\ref{eq:repmix}):
\begin{equation}
    \mathrm{LocalAggr}(X) = X + \mathrm{BN}\big(\mathrm{DW}_{3\times 3}(X) + \mathrm{DW}_{1\times 1}(X)\big),
    \label{eq:repmix}
\end{equation}
where $\mathrm{DW}$ and $\mathrm{BN}$ denote depthwise convolution and batch normalization, respectively. Following~\cite{liu2023efficientvit, yun2024shvit, vasu2023fastvit}, the Local Aggregation can be re-parameterized as a single $3\times3$ depthwise convolution during inference:
\begin{align}
    \mathrm{LocalAggr}(X) = \mathrm{DW}_{3\times 3}(X).
    \label{eq:rep_cpa}
\end{align}
The aggregated features are then fed into two distinct branches: SCA and SSHA. The outputs from the two branches interact through a learnable cross-branch projection, where $W_p$ is a learnable weight that projects the features back into the original channel dimension $C$. This fusion ensures that local and global representations interact coherently. The overall CoDA module is described in Eq.~(\ref{eq:coda_la}) to Eq.~(\ref{eq:coda_proj}).
\begin{align}
    X &= \mathrm{LocalAggr}(X), \label{eq:coda_la} \\
    X_{sca} &= \mathrm{SCA}(X), \label{eq:coda_sca} \\
    X_{att} &= \mathrm{SSHA}(X), \label{eq:coda_ssha} \\
    X &= W_p * [X_{sca},\; X_{att}]. \label{eq:coda_proj}
\end{align}
\subsubsection{Spatial Convolutional Attention (SCA)}
The first branch applies Spatial Convolutional Attention to model fine-grained local details. SCA computes both max-pooled and average-pooled spatial descriptors, then processes the result through a $3\times3$ convolution followed by a sigmoid activation:
\begin{align}
    \mathcal{M}_{a} &= \sigma\Big(\mathrm{Conv}_{3\times3}\big([\mathrm{max}(X),\;\mathrm{avg}(X)]\big)\Big), \\
    \mathrm{SCA}(X) &= X \odot \mathcal{M}_{a}.
    \label{eq:sca}
\end{align}
This operation produces an attention mask $\mathcal{M}_{a} \in \mathbb{R}^{1 \times H \times W}$ that selectively emphasizes salient spatial locations. The output feature map $X_{sca} \in \mathbb{R}^{C \times H \times W}$ from SCA is the Hadamard product of the input features and the attention mask, giving $\mathcal{O}(NC)$ complexity. This linear in token count complexity, enabling SCA to capture local structures often missed by global attention with almost negligible overhead.
\begin{table*}[!ht]
\centering
\caption{Comparison of CoDAT attention mechanism designs with others vision transformers.}
\label{tab:attn_design_comparison}
\setlength{\tabcolsep}{4pt}
\begin{tabular}{m{2cm} ccc c p{9.4cm}}
\hline
\multirow{3}{*}{\textbf{Method}}
  & \multicolumn{3}{c}{\textbf{Attention Properties}}
  & \multirow{3}{*}{\textbf{Complexity}}
  & \multicolumn{1}{c}{\multirow{3}{*}{\textbf{Approach}}} \\
\cline{2-4}
  & \textbf{Spatial} & \textbf{Channel} & \textbf{Local}  
  & & \\
  & \textbf{Red.} & \textbf{Red.} & \textbf{Attn.} 
  & & \\
\hline
EfficientViT~\cite{liu2023efficientvit}
  & \cmark & \cmark & \xmark 
  & $\mathcal{O}(N \cdot w^2 \cdot C/g)$
  & Cascaded group ($g$) attn for channel reduction with windowed ($w$) tokens for spatial reduction and sequential head conditioning. \\
SHViT~\cite{yun2024shvit}
  & \xmark & \cmark & \xmark 
  & $\mathcal{O}(N^2\cdot p_{dim})$
  & Single-head attn on partial channels $p_\text{dim}{<}C$ at full token resolution with remaining channels receive no global context and no spatial compression. \\
S2AFormer~\cite{XuS2AFormer2025}
  & \cmark & \cmark & \xmark 
  & $\mathcal{O}\left(\frac{N^2}{k^2} \cdot C\right)$
  & 1D strip attn along horizontal and vertical axes independently with
    axis-aligned context only. \\
\hline
\textbf{CoDAT (Ours)}
  & \cmark & \cmark & \cmark 
  & $\mathcal{O}\left(M^2\cdot C_v+NC\right)$
  & Jointly reduces channel ($C_v{<}C$) and long token sequences $\big($from $N^2$ to $M^2$, where $M=HW/r_s^2$$\big)$ with local saliency from SCA $\left(\mathcal{O}\left(NC\right)\right)$ in parallel.\\ \hline
\end{tabular}
\end{table*}
\subsubsection{Strided Single-Head Attention (SSHA)}
We introduce SSHA as one branch of the CoDA module to enhance global dependency modeling with minimal computational overhead. Unlike conventional multi-head self-attention (MHSA), which suffers from high memory traffic and redundant head interactions, SSHA employs a single-head formulation with spatially strided projections, significantly reducing computational and memory costs while preserving global contextual reasoning. 

Let $r_s$ denote the stride ratio and $M$ be the number of tokens such that $M = \frac{H}{r_s}\times \frac{W}{r_s}$. Given an input feature map $X \in \mathbb{R}^{C \times H \times W}$, SSHA applies a strided projection module using standard point-wise convolutions to produce Query $Q \in \mathbb{R}^{C_{qk} \times M}$, Key $ K \in \mathbb{R}^{C_{qk} \times M}$, and Value $V \in \mathbb{R}^{C_v \times M}$. Following \cite{yun2024shvit, liu2023efficientvit}, the query and key channels $C_{qk}{=}16$ are kept small and fixed to reduce the computation of the attention score to $\mathcal{O}(M^2 C_{qk})$, while the value channels $C_v{=}\frac{1}{4}C$ retain sufficient capacity for feature aggregation (discussed in Sect.~\ref{sec:s-abl}), giving a total attention cost of $\mathcal{O}(M^2 C_v)$, a reduction of $r_s^4$-times over the standard full-resolution $\mathcal{O}(N^2C)$. The stride values for each stage are detailed in Table~\ref{tab:arch_detail}. This projection emphasizes the most informative channels while reducing spatial activations, serving as an efficient pre-attention filtering stage. SSHA then computes attention map using the scaled dot-product formulation as detailed in Eq.~(\ref{eq:ssha_sp}) to Eq.~(\ref{eq:ssha_prop}).
\begin{align}
    Q, K, V &= \mathrm{StridedProj}(X), \label{eq:ssha_sp} \\
    \mathrm{Attn}(Q, K, V) &= \mathrm{Softmax}\!\bigg(\frac{QK^T}{\sqrt{C_{qk}}}\bigg)V, \label{eq:ssha_sdp} \\
    \mathrm{SSHA}(X) &= \mathrm{Propagate}\big(\mathrm{Attn}(Q, K, V)\big). \label{eq:ssha_prop}
\end{align}

The resulting attended features are spatially propagated back to recover their original resolution through $3\times3$ depth-wise transposed convolution, with kernel size and stride both equal to $r_s$, followed by a re-folding operation. This produces a refined global representation aligned with the original feature map. The use of Local Aggregation as a pre-filtering step and matched kernel--stride values ensures uniform coverage of the output grid and avoids the checkerboard artifacts commonly associated with transposed convolution. The propagated output $X_{att} \in \mathbb{R}^{C_v \times H \times W}$ captures long-range spatial dependencies while maintaining strict computational efficiency, making SSHA particularly suitable for real-time and resource-constrained deployment scenarios.

Fig.~\ref{fig:coda_ssha_bench} and Table~\ref{tab:attn_design_comparison} compare the proposed Collaborative Dual-Attention with several notable self-attention variants from prior work. The vanilla MHSA in Fig.~\ref{fig:coda_ssha_bench}(a) exhibits quadratic complexity $\mathcal{O}(N^2)$ and suffers from non-local representation limitations. The Cascaded Group Attention in Fig.~\ref{fig:coda_ssha_bench}(b) compresses the token sequences in a window; however, it takes longer to reshape the tokens. The Single-Head Attention in Fig.~\ref{fig:coda_ssha_bench}(d) eliminates multi-head redundancy but retains full spatial resolution, while the Strip Self-Attention in Fig.~\ref{fig:coda_ssha_bench}(c) offers limited global context from a single attention map. In contrast, our method jointly reduces the complexity of long token sequences, eliminates multi-head channel redundancy, and achieves complementary global-local attention representations through the CoDA design.

\subsection{CoDAT Backbone Architecture}
CoDAT backbones employ a three-stage pyramid structure with a $16{\times}16$ Convolutional Stem consisting of four $3{\times}3$ convolutions, each with a stride of 2, followed by batch normalization and an activation function, yielding a total spatial downsampling factor of 16 $\left(\frac{H}{16}{\times}\frac{W}{16}\right)$. Unlike conventional four-stage backbones\cite{wang2022pvt, wang2021pyramid, vasu2023fastvit, maaz2022edgenext} that use a $4{\times}4$ patch embedding, this aggressive early downsampling reduces the token count by $16{\times}$ before entering the transformer stages, directly lowering FLOPs, memory access, and energy per inference, while the hierarchical convolutional receptive field compensates for the reduced spatial granularity.

Following the stem, features are processed through three stages of CoDA Blocks as the core computational unit. Each block integrates convolutional local refinement with collaborative dual-attention for global reasoning, arranged in a residual:
\begin{align}
    X &= X + \mathrm{ConvFFN}(X), \\
    X &= X + \mathrm{CoDA}(X), \\
    X &= X + \mathrm{ConvFFN}(X),
    \label{eq:coda_block}
\end{align}
where the first ConvFFN provides local spatial conditioning before attention, CoDA performs collaborative global--local reasoning (detailed in Section~\ref{sec:coda}), and the second ConvFFN refines the attended features. ConvFFN replaces the standard MLP with a convolutional feed-forward operator:
\begin{equation}
    \mathrm{ConvFFN}(X) = W_r\!\big(\sigma(W_e \cdot \mathrm{DW}_{3\times 3}(X))\big),
    \label{eq:convffn}
\end{equation}
where $W_e$ and $W_r$ are pointwise expansion and reduction projections, $\mathrm{DW}_{3\times 3}$ is a depthwise convolution that injects local spatial context, and $\sigma$ denotes the activation function. The three-stage design further reduces overhead compared to four-stage alternatives by minimizing intermediate patch merging layers while maintaining a sufficient representational hierarchy through progressive channel expansion across stages.

\subsection{Low-Cost Temporal Modeling on CoDAT}
We extend the CoDA architecture to spatio-temporal video understanding through low-cost temporal modeling. Specifically, we adopt the temporal shift operation~\cite{lin2019tsm}, denoted $\mathrm{TShift}$, which injects temporal context by permuting a small subset of feature channels across adjacent frames without introducing any parameters.

\begin{figure}[t!]
    \centering
    \includegraphics[width=\linewidth]{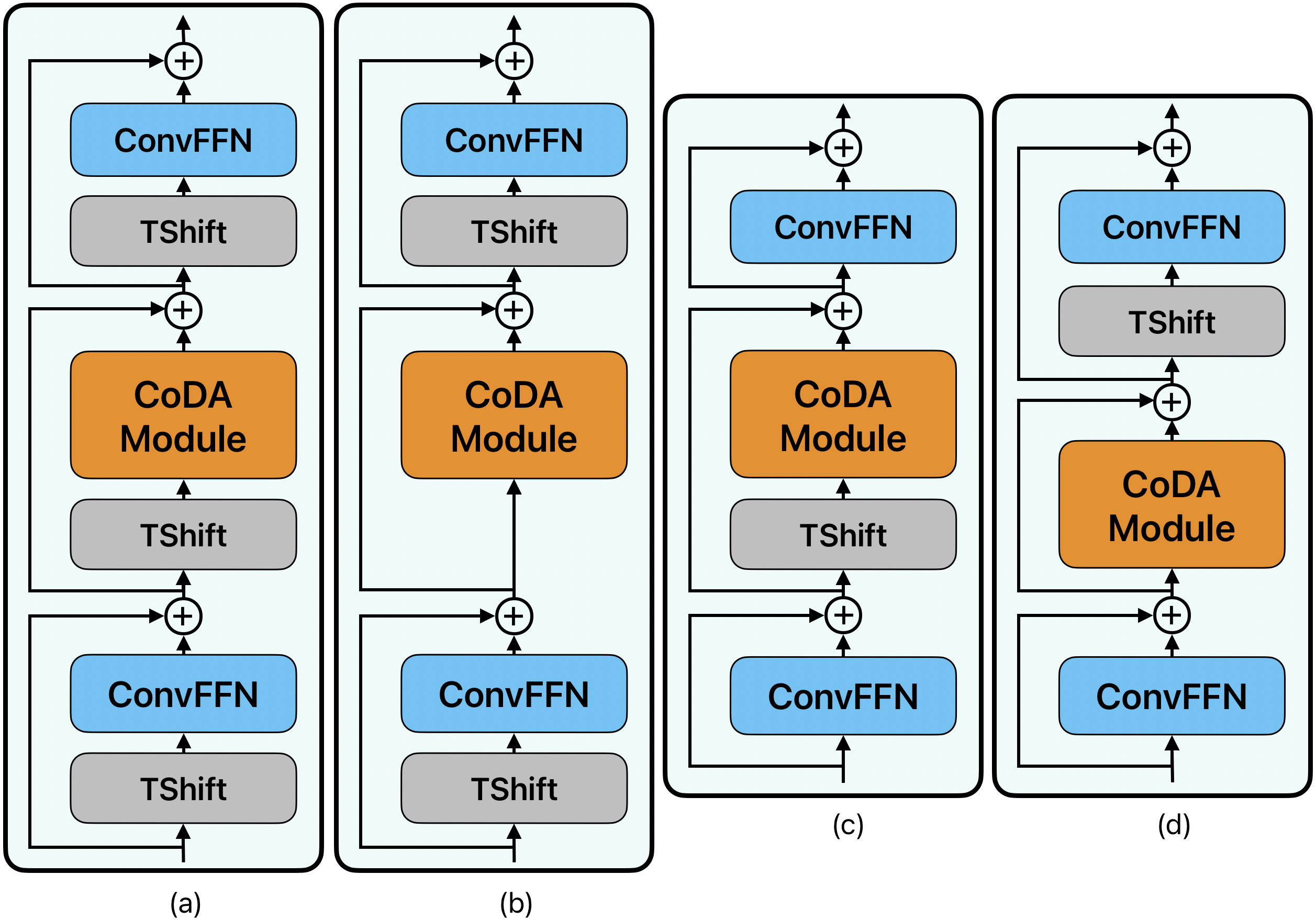}
    \caption{\textbf{CoDAT Temporal Block variants.} (a)~Full-shift A. (b)~Full-shift B. (c)~Attention-shift. and (d)~Final ConvFFN-shift.}
    \label{fig:codat_tsm}
\end{figure}

Given video features $X \in \mathbb{R}^{B\cdot T \times C \times H \times W}$, $\mathrm{TShift}$ will reshaped it into $X \in \mathbb{R}^{B \times T \times C \times H \times W}$ and divides the channels into three groups and shifts them as 
\begin{equation}
    X'_t = \big[\,X_{t-1}^{(c_1)},\; X_{t}^{(c_2)},\; X_{t+1}^{(c_3)}\,\big],
\end{equation}
where $c_1$, $c_2$, and $c_3$ denote the backward-shifted, stationary, and forward-shifted channel subsets, respectively. We insert $\mathrm{TShift}$ immediately in the final ConvFFN layer of each CoDA Block, as detailed in Eq.~\ref{eq:tshift_placement} and Fig.~\ref{fig:codat_tsm}(d):
\begin{align}
    X &= X + \mathrm{ConvFFN}\!\left(\mathrm{TShift}(X)\right).
    \label{eq:tshift_placement}
\end{align}
This placement ensures that CoDA always operates on temporally clean, single-frame (2D) features, after which $\mathrm{TShift}$ injects adjacent-frame context into the resulting global-local representation before ConvFFN integrates both spatial and temporal cues as a final refinement step. Placing $\mathrm{TShift}$ on CoDA would cause SSHA and SCA to receive temporally mixed inputs that conflate features from adjacent frames, violating the within-frame spatial coherence assumed by both branches. Additionally, we evaluate several temporal integration configurations: \emph{full-shift}, in which $\mathrm{TShift}$ is applied to both ConvFFN and CoDA modules, and \emph{attention-shift}, in which $\mathrm{TShift}$ is applied solely to the attention module. All configurations serve as ablation baselines to assess temporal sensitivity.

Based on~\cite{lin2019tsm}, each $\mathrm{TShift}$ layer enlarges the temporal receptive field (TRF) by 2, equivalent to a temporal convolution with a kernel size of 3, giving a total TRF of $2L{+}1$ frames for $L$ stacked layers. CoDAT-S (total depth $L{=}5$, see Table~\ref{tab:arch_detail}) yields a TRF of 11 frames, which exceeds the standard 8-frame input. So every frame, including those at the clip boundary, receives context from at least 6 neighboring frames. For 16-frame inputs, deeper variants CoDAT-M and CoDAT-L provide full temporal coverage, while CoDAT-S covers central frames fully but edge frames partially (See ablation on Table~\ref{tab:t-abl}).
% ---- CoDAT Block Variants Figure ----

\section{Experiment}
In order to assess the effectiveness of CoDAT, comprehensive experiments were conducted. The experiments involve the ImageNet-1K dataset \cite{russakovsky2015imagenet} for image classification, the Kinetics-400\cite{kay2017kinetics} dataset for coarse video action recognition, the MA-52\cite{guo2024benchmarking} dataset for low-intensity micro action recognition, and the UCF-101\cite{soomro2012ucf101} dataset for small-scale “human-object” interactions and sports. Beyond accuracy, deployment efficiency on resource-constrained IoT edge hardware is evaluated through energy consumption and latency, reported per-image for classification and per-frame for action recognition.

\subsection{Edge Energy and Inference Test}
All inference tests on edge hardware use ONNX Runtime. On Jetson AGX Orin, the model executed with CUDA at a maximum power configuration of 50 watts. Power is measured via the \texttt{jtop}, which reads the onboard INA3221 current sensors, reporting total power in milliwatts without external instrumentation. On Raspberry Pi~5, inference uses CPU execution with \texttt{num\_threads=4}. Board power is measured directly from the onboard MXL7704 PMIC via \texttt{vcgencmd pmic\_read\_adc}, which reads per-rail voltage and current from the hardware. Total power is derived by summing $\bar{P} = V \times I$ across all matched voltage-current rail pairs.
For both platforms, each model undergoes a 10-second warm-up before measurements begin. Timed measurements consist of 100 repetitions per run over 3 independent runs, with a cool-down of 30 seconds to prevent thermal throttling. Finally, energy per frame is derived as:
% ---- Architecture Variants Table ----
\begin{table}[t!]
\centering
\setlength{\tabcolsep}{3.0pt} % Wider column gap (default 6pt)
\caption{Architecture details of CoDAT variants}
\begin{tabular}{lccccc}
\hline
\multirow{2}{*}{Variant} & Depth & Dims. & $C_v$ & $r_s$ & FFN \\
       & $[L_1, L_2, L_3]$ & $[C_1, C_2, C_3]$ & ratio & ratio & Exp. \\
\hline
CoDAT-S & $[1,2,2]$  & $[192,384,448]$ & 1/4 & $[2,2,1]$ & 2 \\
CoDAT-M & $[2,4,4]$  & $[200,384,448]$ & 1/4 & $[2,2,1]$ & 2 \\
CoDAT-L & $[5,5,4]$  & $[280,448,512]$ & 1/4 & $[4,2,1]$ & 2 \\
\hline
\end{tabular}
\label{tab:arch_detail}
\end{table}
\begin{equation}
    E_\text{frame} = \frac{\bar{P} \times T_\text{total}}{N_\text{frames}
    \times N_\text{rep}} \quad [\text{mJ/F}]
\end{equation}
where $T_\text{total}$ is the total elapsed inference time, $N_\text{rep}$ is the number of repetitions, and $N_\text{frames}$ is the number of input frames per clip ($N_\text{frames}$ is equal to batch size for image classification test). The latency (ms/F) and energy (mJ/F) values are normalized per frame by dividing the total latency by $N_\text{frames}$, ensuring a consistent per frame basis across all models regardless of resolution or capacity. Multi-view configurations (e.g., $3\,\text{crops} \times 10\,\text{clips}$) are applied exclusively for top-1 accuracy evaluation based on each model default configuration. Latency and energy are always measured under a single-clip, single-crop setting to reflect the true single-inference deployment cost.
\begin{table*}[!ht]
\centering
\setlength{\tabcolsep}{3.0pt} % Wider column gap (default 6pt)
\caption{CoDAT Benchmark with SOTA on ImageNet-1K dataset for Image Recognition.}
\begin{tabular}{llccccccccccccc } \hline
\multirow{3}{*}{Model}& \multirow{3}{*}{Venue} & \multirow{3}{*}{Par} & \multirow{3}{*}{FLOP} & \multirow{3}{*}{Input Res.} & Top-1 & \multicolumn{4}{c}{Jetson-AGX-Orin} & \multicolumn{3}{c}{RPi-5}\\ \cline{7-14}

      &    &   &  &   & Acc  & THP & Lat & Power & Energy & THP & Lat & Power & Energy   \\ 
      &    & (M) & (G) &     & (\%) & (img/s) & ($ms$) & (Watt) & (mJ/img)& (img/s) & ($ms$) & (Watt) & (mJ/img)  \\ \hline
PVTv2-B0\cite{wang2022pvt} & CVMJ 2022 & 3.4 & 0.60 & 224$\times$224 & 70.5 & 502.9 & 1.41 & 15.3 & 21.4 & 21.2 & 46.97 & 7.0 & 317.4 \\
iFormer-T~\cite{zheng2025iformer} & ICLR 2025 & 2.9 & 0.53 & 224$\times$224 & 74.1 & 1017.1 & 0.76 & 13.6 & 10.2  & 43.5 & 23.00 & 7.8 & 180.9\\
EdgeNext-XS~\cite{maaz2022edgenext} & ECCVW 2022 & 2.3 & 0.54 & 224$\times$224 & 75.0 & 574.7 & 1.54 & 13.5 & 10.2 & 22.1 & 44.65 & 7.1 & 306.0 \\
FastViT-T8~\cite{vasu2023fastvit} &ICCV 2023 & 3.7 & 0.70 & 256$\times$256 & 75.6 & 657.6 & 0.99 &  15.0 & 14.6 & 26.6 & 37.34 & 7.5 & 284.4 \\
EfficientFormerv2-S0~\cite{li2023rethinking} &ICCV 2023 & 3.5 & 0.40 & 224$\times$224 & 75.7 & 902.7 & 0.80 & 13.9 & 11.4 & 53.2 & 18.61 & 7.7 & 138.9  \\
S2AFormer-mini~\cite{XuS2AFormer2025} & TIP-2025 & 5.02 & 0.56 & 256$\times$256 & 75.7 & - & - & - & - & - & - & - & -  \\
EfficientViT-M5~\cite{liu2023efficientvit}  &CVPR 2023 & 12.4 & 0.52 & 224$\times$224 & 77.1 & 1234.7 & 1.30 & 14.7 & 19.5 & 71.6 & 13.78 & 8.1 & 118.7 \\
MicroViT-S3~\cite{setyawan2025microvit}    &ISCAS 2025 & 16.7 & 0.58 & 224$\times$224 & 77.1 & 1323.7 & 0.74 & 13.1 & 9.4 & 59.3 & 16.71 & 7.9 & 132.5  \\
SHViT-S3~\cite{yun2024shvit} &CVPR 2024 & 14.2 & 0.60 & 224$\times$224 & 77.4 & 1254.3 & 0.92 & 13.1 & 11.7  & 70.1 & 14.17 & 8.3 & 118.3 \\
\rowcolor{gray!20}
CoDAT-S         & - & 9.9 & 0.58 & 256$\times$256 & 77.6 & 1439.6  & 0.69 & 13.1 & 8.3  & 59.1 & 16.95 & 7.4 & 125.3 \\
\hline
PVTv2-B1\cite{wang2022pvt} & CVMJ 2022 & 13.1 & 2.10 & 224$\times$224 & 78.7 & 293.1 & 2.28 & 15.8 & 36.5 & 6.3 & 116.09 & 7.1 & 826.7 \\
iFormer-S~\cite{zheng2025iformer} & ICLR 2025 & 6.5 & 1.09 & 224$\times$224 & 78.8 & 703.1 & 1.01 & 14.1 & 14.6 & 25.1 & 39.29 & 7.8 & 321.1 \\
EfficientFormerv2-S1~\cite{li2023rethinking} &ICCV 2023 & 6.1 & 0.65 & 224$\times$224 & 79.0 & 902.7 & 1.07 & 14.1 & 15.2 & 36.5 & 26.79 & 7.7 & 204.4  \\
S2AFormer-XS\cite{XuS2AFormer2025} & TIP-2025 & 6.54 & 1.03 & 256$\times$256 & 79.3 & - & - & - & - & - & - & - & - \\
EdgeNext-S~\cite{maaz2022edgenext} & ECCVW 2022 & 5.6 & 1.26 & 224$\times$224 & 79.4 & 413.5 & 1.90 & 14.1 & 20.7 & 13.1 & 75.85 & 7.3 & 543.8 \\
SHViT-S4~\cite{yun2024shvit}            & CVPR 2024 & 16.5 & 0.99 & 256$\times$256 & 79.4 & 1014.0 & 1.16 & 13.5 & 15.7 & 45.1 & 21.94 & 7.9 & 185.9  \\
EfficientViT-M5$_{384}$~\cite{liu2023efficientvit} & CVPR 2023 & 12.4 & 1.49 &  384$\times$384 & 79.8 & 387.7 & 1.29 & 15.9 & 32.5 & 22.7 & 43.47 & 7.9 & 377.7  \\
FastViT-S12~\cite{vasu2023fastvit}         &ICCV 2023 & 8.8 & 1.82 & 256$\times$256 & 79.8 & 416.1 & 1.35 & 16.0 & 21.2 & 12.8 & 72.57 & 7.7 & 544.9   \\
\rowcolor{gray!20}
CoDAT-M & - & 17.1 & 0.94 & 256$\times$256 & 79.7 & 1021.9 & 1.15 & 13.2 & 15.9 & 38.5 & 25.97 & 8.3 & 201.1 \\
\hline
% iFormer-M       & ICLR 2025 & 8.9 & 1.64  & 224$\times$224  & 80.4 & 542.5 & 16.42 & \\
FastViT-SA12~\cite{vasu2023fastvit}  & ICCV 2023 & 10.9 & 1.95 & 256$\times$256 & 80.6 & 402.4 & 1.41 & 16.1 & 22.7 & 12.8 & 78.18 & 7.9 & 588.4 \\
SHViT-S4$_{384}$~\cite{yun2024shvit} & CVPR 2024 & 16.5 & 2.23 & 384$\times$384 & 81.0 & 485.0 & 1.63 & 13.6 & 22.2 & 19.2 & 51.77 & 8.1 & 410.8  \\
Swin-T~\cite{liu2021swin} & ICCV 2021 & 28.3 & 4.38 & 224$\times$224 & 81.3 & 218.7 & 3.12 & 16.0 & 50.3 & 5.9 & 166.52 & 7.9 & 1388.6 \\
ViT-S/DeiT-S~\cite{touvron2021training} & ICML 2021 & 87.0 & 17.60 & 224$\times$224 & 81.8 & 183.5 & 1.71 & 16.1 & 29.7 & 7.9 & 126.12 & 8.3 & 1103.2 \\
\rowcolor{gray!20}
CoDAT-L & - & 27.8 & 2.19 & 256$\times$256 & 81.4 & 517.9 & 1.33 & 17.0 & 22.6 & 17.2 & 57.57 & 8.0 & 463.3 \\
\hline
\end{tabular}
\label{tab:imnet_1k}
\end{table*}
\subsection{Image Classification}
\subsubsection{Dataset and Training Settings} Image classification experiments are conducted using the ImageNet-1K dataset, which consists of 1.28 million training images and 50,000 validation images across 1,000 categories. Our study follows the training recipe and data augmentation from \cite{yun2024shvit, liu2023efficientvit} with a total of 300 epochs at a resolution of 256$\times$256. It employed the AdamW~\cite{you2019large} with a base learning rate (lr) of $10^{-3}$ and a total batch size of 2048 on 4$\times$4090 GPUs for most CoDAT backbone models. The performance evaluation was conducted without any pre-training or knowledge distillation. We present their throughput for inputs with a batch size of 64 and a single batch size for latency using ONNX Runtime. Following ~\cite{vasu2023fastvit, yun2024shvit, zheng2025iformer}, the batch normalization layers of the CoDAT backbone model are integrated with their adjacent layers.
\subsubsection{ImageNet-1K Classification Result}
Table~\ref{tab:imnet_1k} reports Top-1 accuracy and inference metrics on Jetson AGX Orin and Raspberry Pi~5. We analyze the results from an accuracy–inference efficiency perspective, where a model is considered superior if it achieves higher accuracy at lower latency or higher throughput.

Among all compared lightweight models, CoDAT-S attains the highest accuracy with a fast inference profile in this tier. On Jetson AGX Orin, it simultaneously achieves the highest accuracy, highest throughput, and lowest latency within its tier, strictly dominating iFormer-T, SHViT-S3, and EfficientViT-M5. Compared with iFormer-T, CoDAT-S improves accuracy by 3.5\% while increasing throughput by 1.42$\times$. Relative to SHViT-S3, it provides 0.2\% higher accuracy with 25.0\% lower latency. Against EfficientViT-M5, CoDAT-S improves accuracy by 0.5\% while reducing latency by 46.8\%. These dominance relationships persist on Raspberry Pi~5, where CoDAT-S maintains the highest accuracy with competitive or superior inference characteristics, confirming a consistent advantage across heterogeneous hardware.

In the medium-capacity regime, CoDAT-M demonstrates competitive accuracy while substantially improving inference efficiency. On Jetson, it delivers 2.64$\times$ higher throughput than EfficientViT-M5$_{384}$ and 2.46$\times$ higher throughput than FastViT-S12, with only 0.1\% lower accuracy than both. Importantly, these baselines require significantly higher latency for comparable recognition performance, indicating that CoDAT-M shifts the efficiency frontier toward faster operating points without sacrificing accuracy. On Raspberry Pi~5, the advantage becomes more pronounced; CoDAT-M achieves 3$\times$ higher throughput than FastViT-S12 and 1.69$\times$ higher throughput than EfficientViT-M5$_{384}$, while also reducing latency by 64.2\% and 40.2\%, respectively. Compared with SHViT-S4, CoDAT-M offers 0.3\% higher accuracy at a similar inference cost, reinforcing its favorable scaling behavior.

At a large scale, CoDAT-L surpasses Swin-T by 0.1\% accuracy while achieving 2.37$\times$ higher throughput and 57.4\% lower latency on the Jetson device, thus strictly dominating it in the accuracy–latency space. Relative to ViT-S/DeiT-S, CoDAT-L delivers 2.82$\times$ higher throughput with only 0.4\% lower accuracy and approximately 3$\times$ fewer parameters, representing a substantially more efficient operating point. Similar trends are observed on Raspberry Pi~5, where CoDAT-L consistently shifts the accuracy–latency frontier toward lower-latency regions compared with Swin-T and DeiT-S. Across all three scales and both hardware platforms, the performance scaling of the CoDAT family follows a smooth progression in accuracy while maintaining favorable inference characteristics, indicating improved computational utilization rather than brute-force capacity expansion. 
\begin{table*}[!ht]
\centering
\setlength{\tabcolsep}{4.0pt} % Wider column gap (default 6pt)
\caption{Comparison Between CoDAT and Several of The SOTA on Kinetics-400 dataset.}
\begin{tabular}{lllcccccccccc } \hline
\multirow{3}{*}{Type} & \multirow{3}{*}{Model} & \multirow{2}{*}{Backbone} & Input &  &  & Top-1 &  \multicolumn{3}{c}{Jetson-AGX-Orin} & \multicolumn{2}{c}{RPi-5} \\ \cline{8-13}
    &   &   & Frame     & Param &  GFLOPs       &  Acc & Latency & Power & Energy & Latency & Power & Energy \\ 
    &   &   & $\times$Res. & (M) & $\times$View & (\%) & (ms/F) & (Watt) & (mJ/F) & (ms/F)  & (Watt) & (mJ/F) \\ \hline
\multirow{6}{*}{CNN} & TSN-MbV2\cite{lin2022tsm} & MobileNetV2 & 8$\times224^2$ & 2.8 & 3.9$\times$1$\times$10 & 66.5 & 1.32 & 28.6 & 37.86 & 17.37 & 6.98 & 121.2 \\
                     & TSM-MbV2\cite{lin2022tsm} & MobileNetV2 & 8$\times224^2$ & 2.8 & 3.9$\times$1$\times$10 & 69.5 & 1.49 & 27.4 & 40.75 & 17.96 & 7.17 & 128.7 \\
                     & I3D\cite{carreira2017quo}& InceptionV1 & 250$\times224^2$ & 12.0 & 108$\times$3$\times$10 & 71.1 & - & - & - & - & - & - \\
                    & MoViNet-A1\cite{lin2022tsm} & MoViNet & 50$\times224^2$ & 4.6 & 6.02$\times$1$\times$1 & 72.1 & 1.36 & 37.5 & 86.84 & 23.61 & 6.50 & 153.4 \\
                     & TSM-R50\cite{lin2022tsm}  & ResNet-50 & 8$\times224^2$ & 24.3 & 33.0$\times$1$\times$10 & 74.1 & 3.91 & 41.1 & 160.89 & 121.4 & 7.67 & 930.9 \\
                    & MoViNet-A2\cite{kondratyuk2021movinets} & MoViNet & 50$\times224^2$ & 4.8 & 10.3$\times$1$\times$1 & 75.0 & 3.12 & 37.9 & 118.36 & 30.39 & 6.69 & 203.4 \\
\hline
\rowcolor{gray!20}
\cellcolor{white} & CoDAT-S & CoDAT-S & 8$\times256^2$ & 10.6 & 4.6$\times$1$\times$10 & 73.3 & 0.89 & 15.5 & 13.79 & 17.77 & 7.62 & 135.6  \\ \rowcolor{gray!20}
\cellcolor{white} & CoDAT-S & CoDAT-S & 8$\times256^2$ & 10.6 & 4.6$\times$3$\times$4 & 73.6 & 0.89 & 15.5 & 13.79 & 17.77 & 7.62 & 135.6 \\ \rowcolor{gray!20}
\cellcolor{white} & CoDAT-M & CoDAT-M & 8$\times256^2$ & 17.7 & 7.5$\times$1$\times$10 & 74.7 & 1.43 & 25.6 & 37.14 & 27.01 & 7.63 & 205.7 \\ \rowcolor{gray!20}
\cellcolor{white}\multirow{-4}{*}{Trans.} & CoDAT-M & CoDAT-M & 8$\times256^2$ & 17.7 & 7.5$\times$3$\times$10 & 75.3 & 1.43 & 25.6 & 37.14 & 27.01 & 7.63 & 205.7 \\ 
\hline
\multirow{7}{*}{CNN} & GC-TSM\cite{hao2022group} & ResNet-50 & 8$\times224^2$ & 25.6 & 33.0$\times$1$\times$10 & 75.4 & 10.83 & 36.4 & 398.84 & 208.97 & 7.0 & 1443.4 \\
                     % & GSF\cite{sudhakaran2023gate}& InceptionV3 & 16$\times224^2$ & 22.2 & 54$\times$3$\times$10 & 75.6 & - & - & - & - & -\\
                     & STM\cite{wang2022learning} & ResNet-50 & 16$\times224^2$ & N/A & 66.5$\times$3$\times$10 & 76.9 & - & - & - & - & - & - \\
                     & STANet\cite{li2023spatio} & ResNet-50 & 16$\times224^2$ & N/A & N/A & 76.4 & - & - & - & - & - & -\\
                     & AGPN\cite{chen2023agpn}& ResNet-50 & 8$\times224^2$ & 27.6 & N/A$\times$3$\times$10 & 76.7 & - & - & - & - & - & - \\
                    & SlowFast 8$\times$8 \cite{feichtenhofer2019slowfast}& ResNet-50 & 32$\times256^2$ & N/A & 65.7$\times$3$\times$10 & 77.0 & 7.05 & 40.9 & 286.33 & 50.56 & 7.48 & 378.3 \\
                    & MoViNet-A3\cite{kondratyuk2021movinets} & MoViNet & 50$\times224^2$ & 5.3 & 56.9$\times$1$\times$1 & 78.2 & 4.04 & 38.3 & 155.37 & 40.06 & 6.74 & 270.0 \\
\hline
\multirow{15}{*}{Trans.} & MViT-S~\cite{fan2021multiscale}  & MViT-S & 16$\times224^2$ & 26.1 & 32.9$\times$1$\times$5 & 76.0 & - & - & - & - & - & - \\
                        & LAPS~\cite{zhang2022long} & Visformer & 8$\times224^2$  & 39.8 & 40.1$\times$3$\times$5 & 76.4 & 9.34 & 43.3 & 410.10 & 148.90 & 7.77 & 1157.3  \\
                        & TokShift~\cite{zhang2021tokenshift} & ViT-B & 8$\times224^2$  & 85.9 & 135$\times$3$\times$10 & 77.3 & 10.15 & 49.9 & 511.35 & 460.64 & 8.80 & 4052.5 \\
                        % & TokShift~\cite{zhang2021tokenshift} & ViT-B & 16$\times224^2$  & 85.9 & 270$\times$3$\times$10 & 78.2 & 10.15 & 49.9 & 511.35 & 509.09 & 3516.4 \\
                        & ViT-Shift~\cite{zhang2024temporal} & ViT-B & 8$\times224^2$ & 85.9 & 135$\times$3$\times$1 & 77.1 & 10.39 & 49.9 & 512.44 & 510.20 & 8.82 & 4089.5 \\
                        & ViT-Shift~\cite{zhang2024temporal} & ViT-B & 8$\times224^2$ & 85.9 & 135$\times$3$\times$10 & 77.6 & 10.39 & 49.9 & 512.44 & 510.20 & 8.82 & 4089.5 \\
                        & TimeSFormer~\cite{bertasius2021space}& ViT-B  & 8$\times224^2$  & 121.4 & 590$\times$3$\times$1 & 78.0 & 10.27 & 54.1 & 591.90 & 600.72 & 8.46 & 5007.1 \\
                        & MViT-B~\cite{fan2021multiscale}  & MViT-S & 16$\times224^2$ & 36.6 & 70.5$\times$1$\times$5 & 78.4 & - & - & - & - & - & - \\
                        & VTN~\cite{neimark2021video} & ViT-B & 16$\times224^2$  & 114.0 & 270$\times$3$\times$10 & 78.6 & - & - & - & - & - & - \\
                        & VSwin-T~\cite{liu2022video}  & Swin-T & 32$\times224^2$ & 28.0 & 88.0$\times$3$\times$4 & 78.8 & 6.96 & 51.7 & 365.20 & 125.16 & 9.09 & 2074.1 \\
                        & UniFormer-S\cite{li2023uniformer} & UniFormer & 16$\times224^2$ & 21.4 & 41.8$\times$1$\times$1 & 78.4 & 6.53 & 41.4 & 268.43 & 195.30 & 7.94 & 1524.5 \\ 
 \rowcolor{gray!20}
\cellcolor{white} & CoDAT-M$_{384}$ & CoDAT-M & 8$\times384^2$ & 17.7 & 16.8$\times$3$\times$4 & 77.3 & 2.56 & 33.9 & 86.8 & 62.70 & 7.96 & 499.0 \\ \rowcolor{gray!20}
\cellcolor{white} & CoDAT-M$_{384}$ & CoDAT-M & 8$\times384^2$ & 17.7 & 16.8$\times$3$\times$10 & 77.5 & 2.56 & 33.9 & 86.8 & 62.70 & 7.96 & 499.0 \\ \rowcolor{gray!20}
\cellcolor{white}& CoDAT-L$_{384}$ & CoDAT-L & 8$\times384^2$ & 28.4 & 39.1$\times$3$\times$4 & 78.5 & 4.74 & 41.9 & 198.7 & 140.34 & 8.04 & 1128.5 \\ \rowcolor{gray!20}
\cellcolor{white}& CoDAT-L$_{384}$ & CoDAT-L & 8$\times384^2$ & 28.4 & 39.1$\times$3$\times$10 & 78.6 & 4.74 & 41.9 & 198.7 & 140.34 & 8.04 & 1128.5 \\
\hline
\end{tabular}
\label{tab:kinetics_400}
\end{table*}
\subsubsection{Energy Efficiency in Image Classification}
Energy per image provides a direct measure of deployment sustainability, and the CoDAT family consistently resides on superior accuracy–energy curves across both hardware platforms. At the small scale on Jetson AGX Orin, CoDAT-S achieves 8.3 mJ/img, reducing energy by 29.1\% compared with SHViT-S3 and by 57.4\% relative to EfficientViT-M5, while delivering higher accuracy. At the medium scale, CoDAT-M consumes 15.9 mJ/img, corresponding to a 51.1\% reduction compared with EfficientViT-M5$_{384}$ and a 25.0\% reduction relative to FastViT-S12, again at competitive accuracy. At the large scale, CoDAT-L requires only 22.6 mJ/img, yielding a 55.1\% reduction versus Swin-T and a 23.9\% reduction relative to DeiT-S while matching or exceeding their recognition performance.

Similar trends hold on Raspberry Pi~5, where CoDAT-M reduces energy by 63.1\% compared with FastViT-S12 and by 46.75\% relative to EfficientViT-M5$_{384}$, and CoDAT-L lowers energy by 66.6\% compared with Swin-T and 58.0\% compared with DeiT-S. Importantly, energy growth across CoDAT variants scales sub-linearly with model capacity, indicating improved computational utilization rather than proportional power inflation. These consistent double-digit to over 50\% reductions across scales and platforms confirm that CoDAT establishes a favorable accuracy–energy operating frontier, making it particularly well-suited for sustained edge deployment under strict power budgets, including the human action understanding tasks detailed in Section~\ref{sec:var-kinetics}.

\subsection{Action Recognition Task}
\label{sec:var-kinetics}
\subsubsection{Dataset and Experimental Settings}
Next, we extend the benchmark performance of CoDAT on three complementary video benchmarks under various configurations and compare it with representative CNN-based, Transformer-based, and hybrid video action recognition methods.

The first experiment is on Kinetics-400~\cite{kay2017kinetics}, comprising 10-second clips of 400 diverse human actions. It captures both human-object and human-human interactions in natural, unconstrained environments, making it a valuable dataset for evaluating spatio-temporal generalization in real-world video understanding tasks. For Kinetics-400, we apply AdamW with $lr=2e^{-4}$, 5 epochs for warm-up, and train for 70 epochs with a batch size of 256 on each 4$\times$GPUs. Similar to the image classification task, several data augmentations are applied during training, such as Mixup, CutMix, auto-augmentation, random erasing, and label smoothing to prevent overfitting.

The next video benchmark is on MA-52~\cite{guo2024benchmarking}, a specialized video dataset designed for fine-grained micro action recognition across 52 domain-specific categories. It features short video clips with subtle temporal variations, emphasizing detailed motion cues and context-dependent behaviors. The dataset is well-suited for evaluating models under strict temporal sensitivity, particularly in scenarios where frame-level precision and nuanced visual understanding are critical. For MA-52, we applied a similar training recipe as with Kinetics-400, except for 80 epochs and a total batch size of 16.

The other benchmark is on UCF-101~\cite{soomro2012ucf101}, a widely adopted small-scale dataset of 13,320 clips across 101 action categories that emphasize human-object interactions (e.g., playing instruments, applying makeup) and repetitive body motions (e.g., push-ups, jumping jacks) under a controlled background, making it a standard testbed for evaluating transfer learning and fine-tuning efficiency on downstream action recognition tasks. We initialize CoDAT with both ImageNet-1K and Kinetics-400 pre-trained weights and fine-tune using AdamW with a learning rate of $2e^{-4}$, 5 warm-up epochs, and a total of 50 training epochs with a batch size of 32 on each of 4$\times$ GPUs. The same augmentation strategy used for Kinetics-400 is retained. Following the standard evaluation protocol, we report the average Top-1 accuracy over the official train/test splits~1. We also use UCF-101 for the model robustness test by compressing it with an MPEG-4 video decoder with a GoP size of 12. This compressed and small-scale transfer setting  complements the large-scale Kinetics-400 and fine-grained MA-52 evaluations by assessing whether CoDAT's learned representations generalize effectively to domain-specific, data-limited, and compressed scenarios, which is a practical requirement for edge deployment, where task-specific labeled data is often scarce.

\subsubsection{Results on Kinetics-400 Action Recognition Dataset}
As shown in Table~\ref{tab:kinetics_400}, compared to lightweight CNN-based models such as TSN-MbV2 and TSM-MbV2, CoDAT-S achieves a Top-1 accuracy of 73.3\%, representing a significant improvement of 3.8\% over TSM-MbV2 while also delivering lower latency in similar test scenarios. When extended to multi-view testing with 3 crops $\times$ 4 clips, CoDAT-S further improves to 73.6\% Top-1, outperforms I3D\cite{carreira2017quo}, and matches the accuracy of heavier CNN models such as TSM-R50 and GC-TSM while being 4.5$\times$ faster than GC-TSM. Moreover, the CoDAT-M$_{384}$ can outperform the CNN-based video recognition SOTA such as STM~\cite{wang2022learning}, STANet~\cite{li2023spatio},  AGPN~\cite{chen2023agpn}, and SlowFast~\cite{feichtenhofer2019slowfast} with more than 2$\times$ fewer GFLOPs and 2.75$\times$ faster than SlowFast 8$\times$8 \cite{feichtenhofer2019slowfast}. This highlights the efficiency of the CoDAT in temporal tasks.

Compared to Transformers using Temporal-shift-based methods like LAPS~\cite{zhang2022long}, TokShift~\cite{zhang2021tokenshift}, and ViT-Shift~\cite{zhang2024temporal}, CoDAT-M$_{384}$ achieves 0.9\% higher Top-1 accuracy than LAPS. While the test scenario is similar to 8 frames, 3 spatial crops, and 10 clips of temporal sampling, CoDAT-M$_{384}$ matches the performance of ViT-Shift and outperforms TokShift with only 16.8 GFLOPs, resulting in a speed increase of 3.6$\times$ to 3.9$\times$ faster than LAPS and TokShift, respectively. At the higher CoDAT variant, CoDAT-L$_{384}$ achieves comparable accuracy with VTN~\cite{neimark2021video} and VSwin-T~\cite{liu2022video} while being $2.9\times$ faster than VSwin-T~\cite{liu2022video}.
\begin{table}[t!]
\centering
\setlength{\tabcolsep}{5pt} % Wider column gap (default 6pt)
\caption{Comparison Between CoDAT and SOTA on MA-52 dataset.}
\begin{tabular}{llcccccc } \hline
\multirow{2}{*}{Type} & \multirow{2}{*}{Model} & Input & \multicolumn{2}{c}{Top-1 (\%)} & Lat \\ \cline{4-5}
        &         & Res.  & Coarse & Fine & (ms)  \\ \hline

\multirow{5}{*}{CNN} & TSM-R50~\cite{lin2022tsm}  & 8$\times224^2$ & 77.64 & 56.75 & 3.8 \\
                     & SlowFast 8 \cite{feichtenhofer2019slowfast} & 32$\times256^2$ & 77.18 & 59.60 & 6.7 \\
                     & I3D-R50\cite{carreira2017quo} & 64$\times225^2$  & 78.16 & 57.07 & 6.3 \\
                     & MANet\cite{guo2024benchmarking}  & 8$\times224^2$ & 78.95 & 61.33 & 3.9\\
                     & DualActNet\cite{yu2024dualactnet} & 32$\times224^2$ & 80.14 & 61.86 & - \\
\hline
\multirow{4}{*}{Trans.} & TimeSFormer~\cite{bertasius2021space}  & 8$\times224^2$ & 69.17 & 40.67 & 10.1 \\
                        & VSwin-T~\cite{liu2022video}  & 32$\times224^2$ & 77.95 & 57.23 & 6.8\\
                        & UniFormer~\cite{li2023uniformer} & 16$\times224^2$ & 79.03 & 58.89 & 12.5\\             
\rowcolor{gray!20}
\cellcolor{white} & CoDAT-M$_{384}$ & 8$\times384^2$ & 79.18 & 63.19 & 2.5 \\
\hline
\multicolumn{6}{l}{*Latency reported from Jetson-AGX Orin}
\end{tabular}
\label{tab:ma_52}
\end{table}
\subsubsection{Results on Fine-grained Micro Action Recognition}
Table~\ref{tab:ma_52} reports the benchmark of the temporal sensitivity and fine-grained micro recognition capabilities of CoDAT on the Micro Action-52 (MA-52) dataset. Among CNN models, MANet\cite{guo2024benchmarking} and DualActNet\cite{yu2024dualactnet} achieve strong coarse-level Top-1 accuracies of 78.95\% and 80.14\%, respectively. However, their fine-grained recognition performance remains limited by the locality of the convolutional backbone and its reliance on stacked temporal modeling. Notably, CoDAT-M surpasses all CNN baselines with a fine-grained accuracy that is 1.86\% higher than MANet\cite{guo2024benchmarking} and 1.33\% higher than DualActNet\cite{yu2024dualactnet}. In coarse, CoDAT shows competitive accuracy while being 1.56$\times$ times faster than MANet\cite{guo2024benchmarking} and 2.7$\times$ times faster than SlowFast\cite{feichtenhofer2019slowfast}.

Transformer-based models such as TimeSFormer~\cite{bertasius2021space} and VSwin-T~\cite{liu2022video} offer rich spatio-temporal modeling but incur high computational costs and latency. TimeSFormer~\cite{bertasius2021space} achieves only 40.67\% fine accuracy with a latency of 10.1 ms/frame. VSwin-T~\cite{liu2022video} improves to 57.23\% fine accuracy but has a latency of 6.8 ms/frame, making it unsuitable for real-time applications. In contrast, CoDAT-M outperformed VSwin-T~\cite{liu2022video} by 5.96\% while delivering faster inference latency of over 2.7$\times$. Compared to UniFormer-B~\cite{li2023uniformer}, CoDAT-M is 5$\times$ faster while maintaining a 4.3\% accuracy advantage.
\begin{table}[t!]
\centering
\setlength{\tabcolsep}{3.0pt} % Wider column gap (default 6pt)
\caption{Comparison Between CoDAT and SOTA on UCF101 Dataset.}
\resizebox{1.0\linewidth}{!}{
\begin{tabular}{lcccccc} \hline
\multirow{2}{*}{Model} & \multirow{2}{*}{Input} & Par & FLOPs & \multicolumn{2}{c}{Top-1 (\%)} & Lat*  \\ \cline{5-6}
        &        & (M) & (G) &  Orig. & Compr. & (ms/F) \\ \hline
\multicolumn{7}{c}{\textit{Pretrain on ImageNet-1K}} \\ \hline
TLEE-L\cite{wang2023tlee}  & N/A & 34.5 & 24.3 & 84.1 & - & - \\
TDN\cite{wang2021tdn}       & 16$\times224^2$ & 68.8 & 38.5 & 85.8 &- & -\\
TTSN\cite{zhang2023temporal} & 16$\times224^2$ & 68.8 & 80.2 & 86.4 & - & -\\
TTP\cite{huo2019mobile}      & 25$\times224^2$ & 17.5 & 105.2 & - & 87.0 & - \\

CoViAR\cite{wu2018compressed} & 25$\times224^2$ & 83.6 & 361.5 & - & 88.4 & - \\
\hline
TSM-MicroViT-S3~\cite{setyawan2025microvit}  & 8$\times224^2$ & 16.4 & 4.7 & 83.9 & 81.8 & 1.28 \\
TSM-EViT-M5~\cite{liu2023efficientvit}  & 8$\times224^2$ & 12.2 & 4.3 & 86.1 & 85.3 & 1.56 \\
TSM-SHViT-S3~\cite{yun2024shvit}  & 8$\times224^2$ & 14.0 & 4.8 & 86.7 & 85.5 & 1.23 \\
TSM-SHViT-S4~\cite{yun2024shvit}  & 8$\times256^2$ & 16.3 & 7.9 & 87.4 & 86.3 & 1.56 \\
 \rowcolor{gray!20}
CoDAT-S & 8$\times256^2$  & 10.6 & 4.6 & 87.7 & 85.8 & 0.89 \\
\rowcolor{gray!20}
CoDAT-M & 8$\times256^2$  & 17.7 & 7.5 & 88.7 & 87.0 & 1.43 \\
 \hline
\multicolumn{7}{c}{\textit{Pretrain on Kinetics-400}} \\ \hline
I3D\cite{carreira2017quo} & 250$\times224^2$ & 12.0 & 108 & 84.5 & - & - \\
MTRFN\cite{he2022mtrfn}   & 8$\times256^2$  & 51.4 & 90.5 & - & 93.7 & - \\
TSM~\cite{lin2022tsm}    & 8$\times224^2$ & 24.3 & 33.0 & 95.9 & - & 3.91 \\
TokShift~\cite{zhang2021tokenshift}  & 8$\times224^2$  & 85.9 & 135 & 95.4 & - & 10.15 \\
LAPS~\cite{zhang2022long}   & 8$\times224^2$  & 39.8 & 40.1 & 95.4 & - & 9.34\\
\rowcolor{gray!20}
CoDAT-S  & 8$\times256^2$  & 10.6 & 4.6 & 94.2 & 93.5 & 0.89 \\ \rowcolor{gray!20}
CoDAT-M  & 8$\times256^2$ & 17.7 & 7.5 & 95.2 & 94.2 & 1.43  \\ \rowcolor{gray!20}
CoDAT-S  & 8$\times384^2$ & 10.6 & 10.3 & 95.4 & 94.5 &1.60  \\ 
\hline
\multicolumn{7}{l}{*Latency reported from Jetson-AGX Orin with non-compressed video}
\end{tabular}
}
\label{tab:ucf101}
\end{table}
\subsubsection{Results on original and compressed UCF-101}
Table~\ref{tab:ucf101} presents the UCF-101 results under ImageNet-1K and Kinetics-400 pretraining, evaluated on both original RGB and MPEG-4 compressed inputs. With ImageNet-1K pretraining, CoDAT-S achieves 87.7\% Top-1 accuracy, surpassing TDN~\cite{wang2021tdn} and TTSN~\cite{zhang2023temporal} by 1.9\% and 1.3\% while requiring 8.4$\times$ and 17.4$\times$ fewer FLOPs and 6.5$\times$ fewer parameters. Compared to TSM-based lightweight backbones with similar parameters under 20M, CoDAT-S outperforms TSM-MicroViT-S3~\cite{setyawan2025microvit} by 3.8\%, TSM-EfficientViT-M5~\cite{liu2023efficientvit} by 1.6\%, and TSM-SHViT-S3~\cite{yun2024shvit} by 1.0\% with lower latency. Against TSM-SHViT-S4~\cite{yun2024shvit}, which uses a similar input resolution, CoDAT-S still achieves 0.3\% higher accuracy with 1.75$\times$ lower latency. CoDAT-M further widens this advantage, reaching 88.7\% and 87.0\% on original and compressed video, surpassing all lightweight TSM baselines and matching TTP~\cite{huo2019mobile} at 14.0$\times$ fewer FLOPs. 

On K400 pretraining, CoDAT-M achieves 95.2\% on RGB, within 0.7\% of TSM~\cite{lin2019tsm} at 4.4$\times$ fewer FLOPs and 2.7$\times$ lower latency, and within 0.2\% of TokShift~\cite{zhang2021tokenshift} and LAPS~\cite{zhang2022long} while being 7.1$\times$ and 6.5$\times$ faster, with 18.0$\times$ and 5.3$\times$ fewer FLOPs, respectively. At resolution $384^2$, CoDAT-S$_{384}$ matches both TokShift and LAPS at 95.4\% while requiring 13.1$\times$ and 3.9$\times$ fewer FLOPs, 8.1$\times$ and 3.8$\times$ fewer parameters, and running 6.3$\times$ and 5.8$\times$ faster, all at just 1.60\,ms latency per-frame. In the compressed domain, CoDAT-S$_{384}$ surpasses MTRFN~\cite{he2022mtrfn} by 0.8\,pp at 8.8$\times$ fewer FLOPs, confirming that CoDAT matches heavyweight baselines on UCF-101 while maintaining sub-2\,ms edge latency.

\subsubsection{Energy Efficiency Analysis on Action Recognition}
Energy per frame (mJ/F) highlights the deployment sustainability of backbones under real-time inference. On Jetson AGX Orin, CoDAT-S consumes only 13.79 mJ/F, achieving an 84.1\% reduction compared with MoViNet-A1 and a 91.4\% reduction relative to TSM-Res50, while maintaining competitive recognition accuracy. At the medium scale, CoDAT-M requires 37.14 mJ/F, corresponding to an 87.0\% reduction compared with SlowFast 8$\times$8  and a 76.1\% reduction relative to MoViNet-A3. Even against GC-TSM, CoDAT-M lowers energy consumption by over 90\%. 

Compared with Transformer-based methods, the advantage becomes more pronounced. CoDAT-M$_{384}$ reduces energy by 76.2\% relative to VSwim-T, 67.7\% compared with UniFormer-S, and 84.5\% versus TimeSFormer. At the large scale, CoDAT-L$_{384}$ consumes 198.7 mJ/F, yielding a 45.6\% reduction compared with VSwim-T and a 64.4\% reduction relative to TimeSFormer. Similar trends are observed on Raspberry Pi~5, where CoDAT consistently operates at substantially lower energy levels than both CNN and Transformer baselines. These results demonstrate that CoDAT significantly shifts the energy–accuracy frontier toward lower-power operating regimes, enabling high-performance video recognition under strict edge deployment constraints.
\begin{table}[t!]
\centering \footnotesize
\caption{Spatial Ablation of CoDAT-S on ImageNet-1K and UCF-101.}
\resizebox{1.0\linewidth}{!}{
\setlength{\tabcolsep}{1.5pt} % Wider column gap (default 6pt)
\begin{tabular}{lcccccc } \hline
\multirow{2}{*}{Ablation}& Par(M) & GFLOPs & THP & Lat(ms) & Top-1  & Energy\\ 
                        & img / vid & img / vid  & (img/s) &  img / vid & IN / UCF & (mJ/F) \\ \hline
\multicolumn{7}{c}{Architectural Design} \\ \hline
4 Stage  & 6.8 / 7.7 & 0.989 / 8.9 & 704.6 & 1.25 / 4.26  & 78.8 / 88.1 & 155.3 \\
\rowcolor{gray!20}
3 Stage  & 9.9 / 10.6 & 0.578 / 4.6 & 1439.6 & 0.69 / 0.89  & 77.6 / 87.7 & 13.79 \\
\hline
\multicolumn{7}{c}{CoDA Ablation} \\ \hline
\rowcolor{gray!20}
CoDA & 9.9 / 10.6 & 0.578 / 4.6& 1439.6 & 0.69 / 0.89  & 77.6 / 87.7 & 13.79 \\
MDTA\cite{zhang2023afd}    & 11.7 / 12.4 & 0.71 / 5.5 & 1183.8 & 0.75 / 0.99 & 77.9 / 86.5 & 15.93 \\
MHSA\cite{touvron2021training} & 11.7 / 12.3 & 0.668 / 5.6 & 1109.5 & 0.79 / 1.01 & 78.0 / 87.4 & 16.00 \\
w/o SCA  & 9.9 / 10.5 & 0.578 / 4.6 & 1498.2 & 0.81 / 0.83 & 77.2 / 87.0 & 13.36 \\
w/o SSHA  & 9.4 /10.1 & 0.564 / 4.5 & 1521.2 & 0.77 / 0.81 & 76.7 / 85.9 & 13.10 \\
\hline
\multicolumn{7}{c}{SSHA Ablation} \\ \hline
\rowcolor{gray!20}
strided proj & 9.9 / 10.6 & 0.578 / 4.6 & 1439.6 & 0.69 / 0.89  & 77.6 / 87.7 & 13.79 \\
w/o strided & 9.9 / 10.7 & 0.585 / 4.7 & 1381.6 & 0.73 / 0.97 & 77.7 / 87.9 & 15.19 \\
Avg($k=3$) & 9.9 / 10.6 & 0.579 / 4.6 & 1432.4 & 0.70 / 0.96 & 77.6 / 87.7 & 15.15 \\ 
Max($k=3$) & 9.9 / 10.6 & 0.579 / 4.6 & 1430.8 & 0.70 / 0.96 & 77.3 / 85.6 & 15.15\\ 
\hline
\multicolumn{7}{c}{SSHA Ratio Ablation} \\ \hline
$C_v$=1/8 & 9.7 / 10.5 & 0.572 / 4.5 & 1452.3 & 0.68 / 0.88  & 77.2 / 86.9 & 13.69 \\
\rowcolor{gray!20}
$C_v$=1/4 & 9.9 / 10.6 & 0.578 / 4.6& 1439.6 & 0.69 / 0.89  & 77.6 / 87.7 & 13.79 \\
$C_v$=1/2 & 10.3 / 10.9 & 0.587 / 4.7 & 1412.9 & 0.71 / 0.95  & 77.7 / 87.8 & 14.12 \\ 
\hline
\multicolumn{7}{l}{*All tested with 256$\times$256 image resolution, $k$ denotes pooling kernel size}
\end{tabular}
}
\label{tab:s-abl}
\end{table}

\subsection{Ablation Study of CoDAT}
In this part, we examine two stages of ablation, first, we conduct a spatial modeling ablation on CoDAT. Then, in the second stage of ablation, we examine the temporal ablation on CoDAT to determine the best architectural decisions regarding temporal shift integration. 

\subsubsection{Ablation Study on Spatial Design}
\label{sec:s-abl}
Table~\ref{tab:s-abl} reports the spatial ablation results under 256$\times$256 resolution on ImageNet-1K (IN) and UCF-101 (UCF), with deployment metrics measured on the Jetson AGX-Orin.

\textit{Stage Design Impact.} Transitioning from a 4-stage with $4{\times}4$ convolutional patch embedding to a 3-stage with a larger $16{\times}16$ patch architecture reduces image GFLOPs by 41.6\% and video GFLOPs by 48.3\%, which leads to a 104.3\% increase in throughput and a 44.8\% reduction in image latency. More significantly, energy consumption decreases by 91.1\%, while Top-1 accuracy drops by only 1.5\% on IN and 0.45\% on UCF-101. This indicates that the 3-stage configuration, which has low spatial resolution but rich semantic information, improves computational utilization dramatically with minimal accuracy compromise, establishing a substantially better accuracy–efficiency operating point.

\begin{table}[t!]
\centering
\caption{Temporal Ablation on CoDAT-S architecture on UCF-101.}
\resizebox{1.0\linewidth}{!}{
\setlength{\tabcolsep}{2.5pt} % Wider column gap (default 6pt)
\begin{tabular}{lccccccc} \hline
\multirow{2}{*}{Model} & \multirow{2}{*}{Input} & Par & FLOPs & Top-1 & Lat & Pow & Energy \\ 
        &     & (M) & (G) & (\%) & (ms) & (Watt) & (mJ/F)  \\ \hline
% TSN-R50~\cite{lin2022tsm}          & 8$\times224^2$ & 24.3 & 33.0$\times$1$\times$2 & 91.7 & 31.8 \\
TSM~\cite{lin2022tsm}   & 8$\times224^2$ & 24.3 & 33.0 & 95.9 & 3.91 & 41.1 & 160.89 \\
TokShift~\cite{zhang2021tokenshift}& 8$\times224^2$ & 85.9 & 135 & 95.4 & 10.15 & 49.9 & 511.35 \\
\hline
 w/o TShift  & 8$\times256^2$  & 10.6 & 4.6 & 85.8 & 0.83 & 15.5 & 13.10  \\
 Full-Shift A    & 8$\times256^2$  & 10.6 & 4.6 & 86.2 & 1.03 & 15.8 & 15.93 \\
 Full-Shift B    & 8$\times256^2$  & 10.6 & 4.6 & 86.8 & 0.95 & 15.8 & 14.75 \\
 on CoDA    & 8$\times256^2$  & 10.6 & 4.6 & 86.9 & 0.89 & 14.9 & 13.39 \\
 pre-CoDA   & 8$\times256^2$  & 10.6 & 4.6 & 87.1 & 0.89 & 14.9 & 13.39 \\
\rowcolor{gray!20}
 post-CoDA  & 8$\times256^2$  & 10.6 & 4.6 & 87.7 & 0.89 & 14.9 & 13.39 \\
\hline
TShift $1/4$ C & 8$\times256^2$ & 10.6 & 4.6 & 87.6 & 7.2 & 15.1& 14.01   \\
\rowcolor{gray!20}
TShift $1/8$ C  & 8$\times256^2$  & 10.6 & 4.6 & 87.7 & 0.89 & 14.9 & 13.39  \\
TShift $1/16$ C & 8$\times256^2$  & 10.6 & 4.6 & 87.5 & 0.89 & 14.8 & 13.29 \\
\hline
Pretrain-K400 & 8$\times256^2$  & 10.6 & 4.6 & 94.2 & 0.89 & 14.9 & 13.39 \\
Pretrain-K400 & 8$\times384^2$  & 10.6 & 10.3 & 95.4 & 1.60 & 28.6 & 45.60 \\
\hline
CoDAT-S$_{8F}$  & 8$\times256^2$  & 10.6 & 4.6 & 94.2 & 0.89 & 14.9 & 13.39 \\
CoDAT-S$_{16F}$ & 16$\times256^2$  & 10.6 & 9.2 & 94.5 & 0.93 & 20.1 & 18.74 \\
\rowcolor{gray!20}
CoDAT-M$_{8F}$ & 8$\times256^2$  & 17.7 & 7.5 & 95.2 & 1.42 & 25.6 & 36.75 \\
CoDAT-M$_{16F}$ & 16$\times256^2$  & 17.7 & 14.9 & 95.3 & 1.47 & 28.4 & 41.65 \\
\hline
\multicolumn{8}{l}{*Latency \& Energy measured on Jetson AGX Orin using ONNX runtime.}
\end{tabular}
}
\label{tab:t-abl}
\end{table}
\textit{Effectiveness of CoDA Attention.} Replacing CoDA with MDTA, which designs the local global in a sequence, increases parameters by 18.2\% and GFLOPs by 22.8\%, while reducing throughput by 17.8\% and increasing energy by 15.5\%, with only a 0.3\% gain in IN accuracy and a 1.2\% drop in UCF accuracy. Similarly, replacing CoDA with MHSA increases parameters by 18.2\%, decreases throughput by 22.9\%, and raises energy consumption by 16.1\%, while yielding only a 0.4\% ImageNet accuracy gain and a 0.3\% UCF accuracy drop. These results show that CoDA improves energy efficiency by approximately 14–16\% and inference speed by 18–23\% compared to heavier attention mechanisms at comparable accuracy, validating its lightweight spatial modeling design.

\textit{Contribution of SCA and SSHA.} Removing SCA increases throughput by 4.1\% but reduces IN accuracy by 0.5\% and UCF accuracy by 0.8\%, indicating that SCA contributes to stable spatial refinement. Removing SSHA increases throughput by 5.7\% and reduces energy by 5.0\%, but it causes a larger accuracy degradation of 1.2\% (IN) and 2.1\% (UCF), highlighting its stronger role in preserving discriminative spatial representation for video tasks.

\textit{SSHA Design Choices.} Eliminating strided projection reduces throughput by 4.0\% and increases energy by 10.2\%, demonstrating degraded hardware efficiency. Alternative pooling strategies (Avg and Max with $k=3$) slightly reduce IN accuracy by 0–0.4\% and UCF accuracy by up to 2.4\%, while increasing energy by approximately 9.8\%. This confirms that the proposed strided projection design improves both computational locality and deployment efficiency.

\textit{Channel Ratio Sensitivity.} Adjusting the channel ratio $C_v$ from 1/4 to 1/8 improves throughput by 0.9\% and reduces energy by 0.7\%, but decreases IN accuracy by 0.5\% and UCF accuracy by 0.9\%. Increasing to $C_v=1/2$ improves IN accuracy by 0.1\% and UCF accuracy by 0.1\%, but increases energy consumption by 2.4\% and reduces throughput by 1.9\%. The default $C_v=1/4$ therefore provides the most balanced trade-off between recognition accuracy and deployment efficiency.

Overall, the ablation results demonstrate that the proposed designs collectively deliver 40–90\% reductions in computational cost and up to 91\% energy savings with minimal ($\leq$1.5\%) accuracy degradation. These improvements confirm that CoDAT's efficiency gains arise from architectural optimization rather than brute-force parameter scaling.

\subsubsection{Temporal Ablation Study}
\label{sec:Temporal}
Table~\ref{tab:t-abl} presents the temporal design analysis of CoDAT-S on UCF-101 under a unified 8$\times$256$^2$ input setting. We analyze accuracy, latency, and energy from an efficiency–performance trade-off perspective.

\begin{figure*}[!t]
    \centering
    \includegraphics[width=\linewidth]{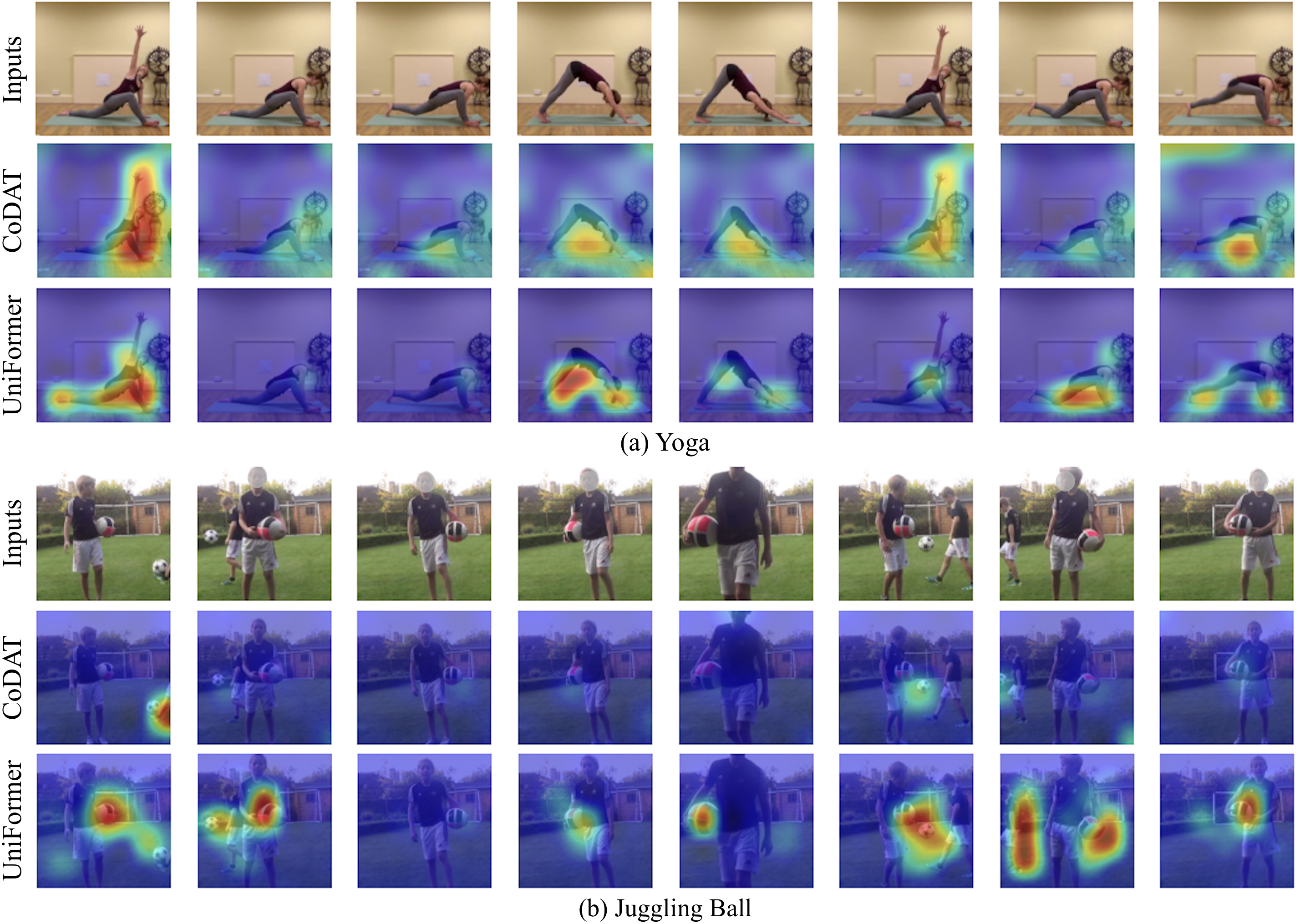}
    \caption{The attention map visualizations on two examples from the Kinetics-400 validation set of CoDAT in comparison with UniFormer. (a): Example on “yoga” category. (b): Example on “juggling ball” category. The first rows of are video frames of inputs, and the second and third rows are the visualizations of CoDAT and UniFormer~\cite{li2023uniformer} attention corresponding to the first row of video frames, respectively.}
    \label{fig:grad_cam}
\end{figure*}

\textit{Effect of $\mathrm{TShift}$ integration on the backbone.} First, we study how temporal modeling should be integrated into the CoDAT, as illustrated in Fig.~\ref{fig:codat_tsm}. We start without $\mathrm{TShift}$, resulting in a Top-1 accuracy of 85.8\%. Then $\mathrm{TShift}$ is integrated into the CoDAT backbone. Among full-shift configurations (Fig.~\ref{fig:codat_tsm}(a) and (b)) and a single temporal shift (Fig.~\ref{fig:codat_tsm}(c) and (d)), a single ConvFFN-$\mathrm{TShift}$ (post-CoDA) in a block achieves the strongest performance, with a Top-1 accuracy of 87.7\% and a latency of only 0.89 ms/frame, improving accuracy by 1.9\% while maintaining only 7.2\% higher latency compared to the no-shift baseline. This result demonstrates that a single $\mathrm{TShift}$ layer in a block is more effective than uniformly shifting features throughout the block, which introduces redundancy without meaningful temporal gain.

\textit{Temporal Placement Strategy.} We further evaluate the temporal shift placement relative to CoDA, as shown in Fig.~\ref{fig:codat_tsm}(c)-(d). Compared to pre-CoDA, on CoDA, and post-CoDA, the $\mathrm{TShift}$ layer in the post-CoDA block yields the highest accuracy among all variants. This behavior indicates that temporal enrichment is most beneficial when introduced after the global-local aggregation within CoDA, enabling the ConvFFN to refine temporally enriched representations and avoid disturbing the attention weights on the features. Thus, this selective configuration is adopted as our proposed CoDAT design, offering the optimal balance between accuracy, latency, and energy hardware efficiency.

\textit{Channel Ratio Sensitivity.} Finally, we vary the fraction of channels participating in the $\mathrm{TShift}$ operation. The small shift ratios, such as $\frac{1}{16}C$ and $\frac{1}{8}C$, yield high accuracy while maintaining low latency. Larger temporal ratios $\left(\frac{1}{4}C\right)$ provide no meaningful accuracy benefits, suggesting that CoDAT requires only minimal temporal displacement to achieve robust motion modeling. These results validate that
temporal cues can be effectively captured even with highly sparse channel shifts, consistent with the lightweight design of the backbone.

\textit{Temporal receptive field and 16-frame capacity.}
To empirically validate the TRF analysis, we evaluate CoDAT-S and CoDAT-M under 16-frame inputs. CoDAT-S$_{16F}$ achieves 94.5\%, a +0.3\% gain over its 8-frame counterpart at only +0.04\,ms/F additional latency despite doubled FLOPs. The modest gain is consistent with CoDAT-S's TRF of 11, which provides full coverage for central frames but partial coverage for the 5 edge frames at each end of a 16-frame clip. CoDAT-M$_{16F}$ shows a smaller improvement of +0.1\% with similarly negligible latency overhead, reflecting that its larger TRF already fully covers all 16 frame positions, leaving little headroom for additional gain. 

\textit{Comparison with TSM and TokShift under K400 Pretraining.} Compared with classical baselines TSM~\cite{lin2022tsm} and TokShift~\cite{zhang2021tokenshift} under a similar Kinetics-400 pretraining strategy, CoDAT-S achieves 94.2\% Top-1 accuracy with only 0.89 ms latency and 13.39 mJ/F energy consumption. Compared with TSM, CoDAT-S exhibits only a 1.7\% accuracy gap while reducing latency by 77.2\% and lowering energy consumption by 91.7\%. Relative to TokShift, CoDAT-S shows a 1.2\% accuracy difference but achieves 91.2\% lower latency  and 97.4\% lower energy. At 384$^2$ resolution, CoDAT-S reaches 95.4\% accuracy, matching TokShift while maintaining 84.2\% lower latency and 91.1\% lower energy. Although TSM achieves slightly higher accuracy, it requires 3.5$\times$ higher latency and over 12$\times$ greater energy consumption. These results indicate that CoDAT-S, especially under K400 pretraining, shifts the accuracy–latency–energy frontier toward dramatically more efficient operating regimes, achieving near-TSM/TokShift accuracy at only a fraction of their deployment cost.

\subsection{Visualization}
Fig.~\ref{fig:grad_cam} illustrates the attention maps generated by CoDAT and compares them with those produced by UniFormer across two representative categories: yoga and juggling a ball on selected Kinetics-400 samples. In the yoga example (Fig.~\ref{fig:grad_cam}(a)), UniFormer~\cite{li2023uniformer} primarily focuses on the upper-body region, especially the raised arm, throughout the sequence. While this region is semantically informative, the attention appears static and narrowly localized. In contrast, CoDAT exhibits a broader and more adaptive attention distribution. It captures the full-body posture and tracks the dynamic transitions across frames, particularly attending to the arms and legs as they shift positions. This dynamic temporal awareness reflects the contribution of the temporal shift in capturing motion-aware features, which complements the efficient spatial modeling of CoDAT with Parallel Attention.

In the juggling ball (Fig.~\ref{fig:grad_cam}(b)), CoDAT again demonstrates the ability to track key object–motion interactions over time. While UniFormer~\cite{li2023uniformer} mostly fixates on the static torso and occasionally on the ball, CoDAT consistently attends to the ball's position across frames, even as it moves. This consistent attention to the object in motion indicates stronger temporal sensitivity and object tracking, which are crucial for distinguishing fine-grained action sequences. The synergy between the temporal shift mechanism and sparse convolution-attention mixing in CoDAT likely contributes to this behavior.

\subsection{Limitations and Future Works}
Although CoDAT offers a strong trade-off between energy efficiency and accuracy on edge devices, its global–local fusion mechanism and spatial–channel compression scheme remain manually designed and fixed. A key future direction is to integrate learnable, content-adaptive token compression that can dynamically choose how many tokens to keep and where to place them across spatial and channel dimensions, following recent work on dynamic token pruning and merging~\cite{kim2024token}. For video analytics, coupling a low-cost $\mathrm{TShift}$ module with the lightweight CoDAT backbone produces a competitive, low-energy, and low-latency action recognition model; however, it is currently constrained to a short and single temporal scale. Future work could extend it to multi-scale and longer-range temporal modeling—e.g., by combining temporal shift with hierarchical temporal pyramids or multi-scale temporal attention~\cite{feichtenhofer2019slowfast, bertasius2021space, li2023uniformer}—to better capture complex and long-duration actions while maintaining the edge-friendly deployment characteristics.

\section{Conclusion}
In this work, we introduce CoDAT, a Collaborative Dual-Attention Transformer that combines Spatial Convolutional Attention with Strided Single-Head Attention to simultaneously capture local detail and global
context under strict compute constraints, augmented by a parameter-free $\mathrm{TShift}$ module for low-cost temporal modeling in multi-frame action recognition. Across ImageNet-1K, Kinetics-400, MA-52, and UCF-101, the CoDAT family consistently delivers a strong accuracy--efficiency balance on edge devices. On ImageNet-1K, CoDAT-M achieves competitive accuracy at roughly $2\times$ higher throughput and lower latency than EfficientViT$_{384}$ and FastViT-S12, while CoDAT-L matches ViT-S accuracy with $3\times$ fewer parameters, $2.8\times$ higher throughput, and $2.5\times$ lower latency. For video action recognition on Kinetics-400, CoDAT matches or exceeds leading CNN, transformer, and hybrid baselines while running $2.9\times$ faster than VSwin-T and nearly $2\times$ faster than ViT-Temporal-Shift models such as LAPS and TokShift, staying within 1\% in accuracy. On MA-52, CoDAT achieves SOTA fine-grained accuracy of 63.6\% while reducing latency by more than $5\times$ and $9\times$ relative to VSwin-T and UniFormer-B, respectively. On UCF-101, CoDAT-S$_{384}$ matches the accuracy of heavyweight ViT-based models at 95.4\% while requiring up to $13\times$ fewer FLOPs and running over $6\times$ faster, confirming effective transfer to small-scale, data-limited scenarios. These results demonstrate that CoDAT provides a scalable, deployment-ready model for distributed IoT perception systems such as smart surveillance, industrial monitoring, and intelligent health care infrastructures.

% \section{Acknowledgement}
%  The authors are grateful for the support of this research. 

\bibliography{ref}
\bibliographystyle{ieeetr}
\vskip -1\baselineskip plus -1fil
\begin{IEEEbiography}[{\includegraphics[width=1in,height=1.25in,clip,keepaspectratio]{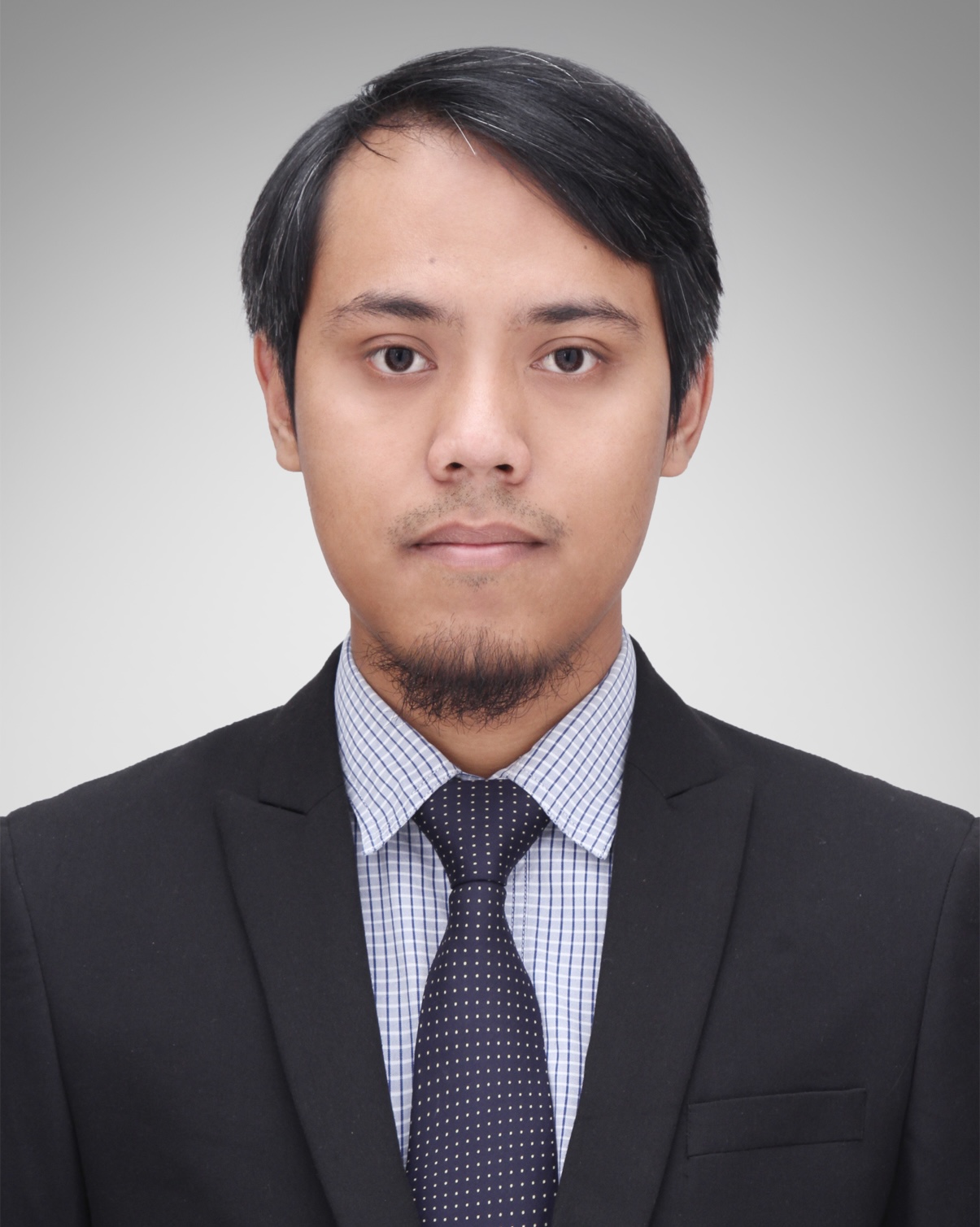}}]{Novendra Setyawan (Student Member, IEEE)}{\space}received the B.Eng. degree in Electrical Engineering from the University of Muhammadiyah Malang, Indonesia, in 2015, and the M.Eng. degree in Electronic Engineering from the Institut Teknologi Sepuluh Nopember (ITS), Surabaya, Indonesia, in 2017. He is currently pursuing the Ph.D. degree in the Department of Electro-Optics Engineering, National Formosa University, Taiwan. He is also a Lecturer in the Department of Electrical Engineering, University of Muhammadiyah Malang, Indonesia. His research has been published in IEEE T-BIOM, IEEE APCCAS, IEEE ISCAS, IEEE ICIP, and other international venues. His research interests include deep learning based image processing, computer vision, and efficient model design for resource-constrained platforms.
\end{IEEEbiography}
\vskip -1\baselineskip plus -1fil
\begin{IEEEbiography}[{\includegraphics[width=1in,height=1.25in,clip,keepaspectratio]{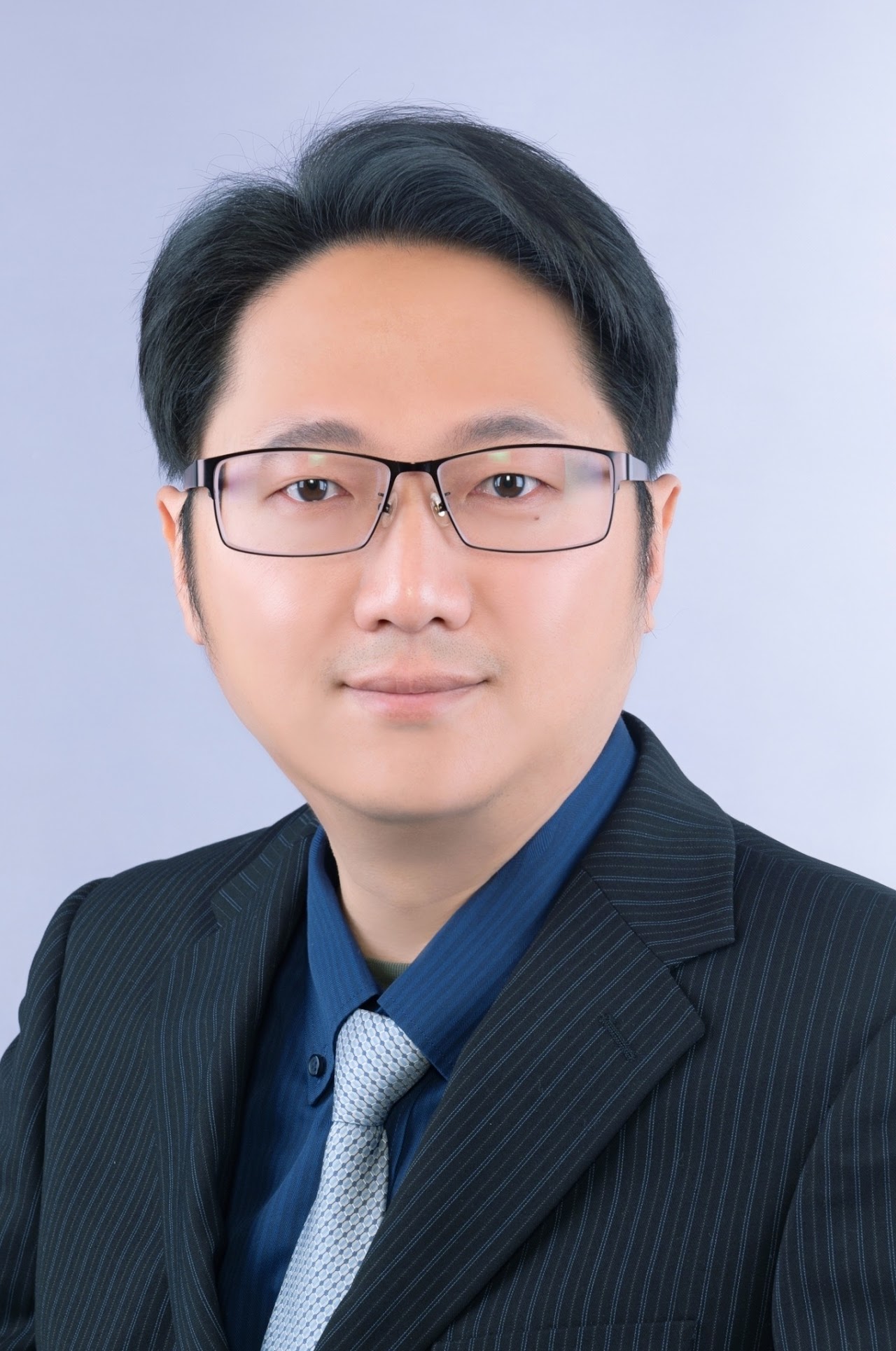}}]{Chi-Chia Sun (Member, IEEE)}{\space} received the B.S. degree in computer science and engineering from National Taiwan Ocean University, Taiwan, in 2004, the M.S. degree in electronic engineering from the National Taiwan University of Science and Technology, Taipei, Taiwan, in 2006, and the Dr.-Ing. degree (with Federal Republic of Germany DAAD Full Scholarship) from Dortmund University of Technology, Germany, in 2011. He was a Principal Engineer with TSMC, specializing in EDA design. He is currently a Full Professor with the Department of Electronic and Computer Engineering, National Taiwan University of Science and Technology. His research interests include image processing, system integration, and VLSI/FPGA design. He is a member of IEEE CASS, VSA-TC and SPS.
\end{IEEEbiography}
\vskip -1\baselineskip plus -1fil
\begin{IEEEbiography}
[{\includegraphics[width=1in,height=1.25in,clip,keepaspectratio]{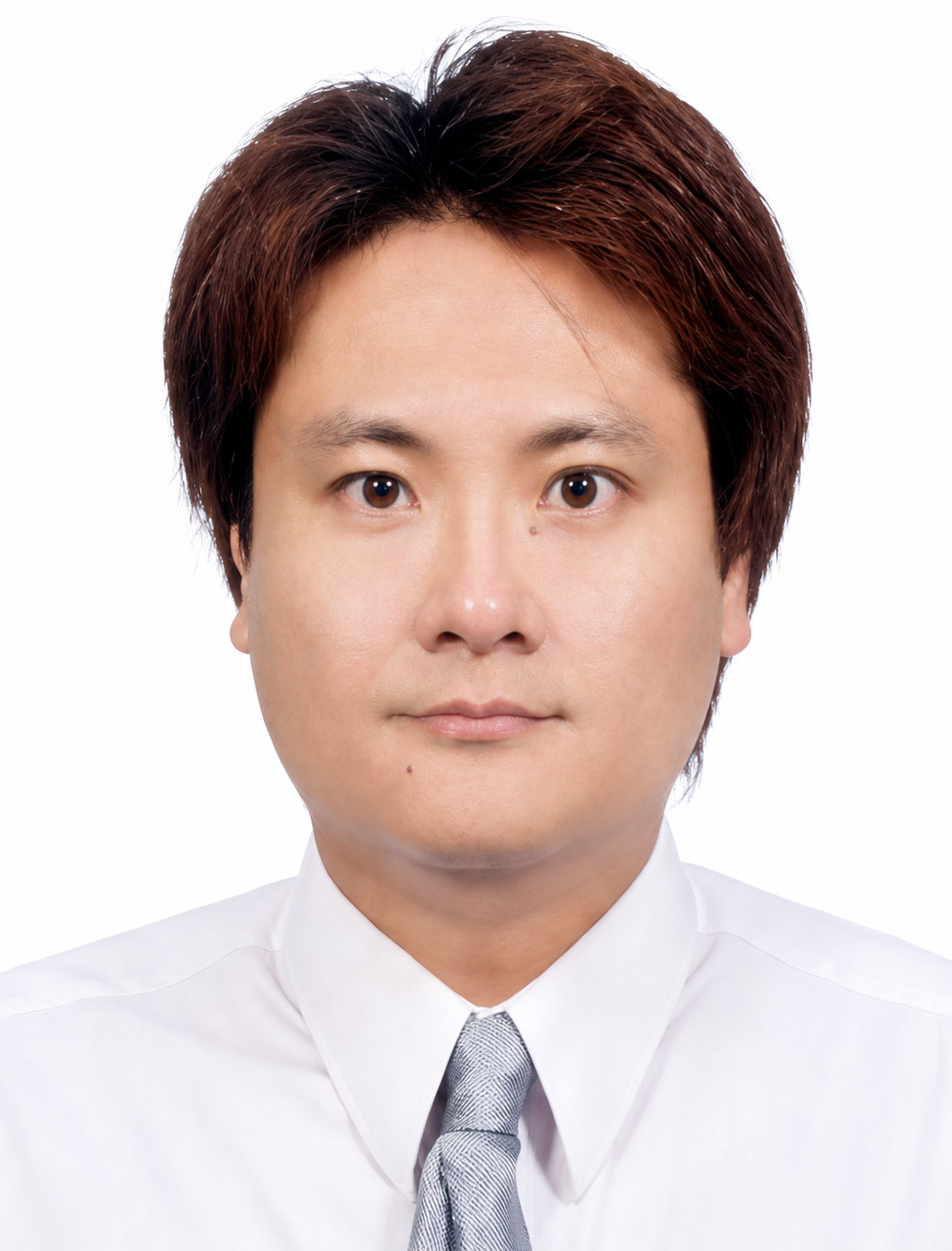}}]{Mao-Hsiu Hsu (Member, IEEE)}{\space} obtained his Ph.D. in electronic and computer engineering from National Taiwan University of Science and Technology in 2005. His career began in 1997 at Motorola, working on crystal filter and oscillator design. He joined Foxconn in California from 2007 to 2012, and in 2013, he was the Inpaq USA Site Country Manager. From 2014 to 2019, he was a Senior Manager at Primax Taiwan. He is currently an Associate Professor in Electro-Optical Engineering at National Formosa University, having previously served as an Assistant Professor in the same department from 2023 to 2025. He possesses extensive expertise in sensor and optical signal processing and deep learning algorithms for fingerprint and facial recognition sensor IC systems, with over 100 technical publications and patents. His current research focuses on sensing systems for Diffusion face ID, fingerprint detection under DeepFake GAN machine learning, and high-speed SerDes re-timer and Tiny ML IC design.   
\end{IEEEbiography}
\vskip -1\baselineskip plus -1fil
\begin{IEEEbiography}[{\includegraphics[width=1in,height=1.25in,clip,keepaspectratio]{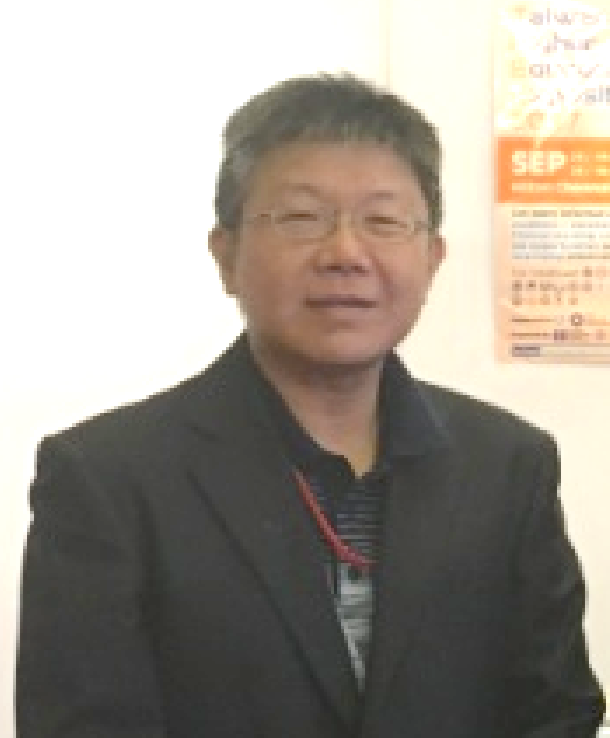}}]{Wen-Kai Kuo (Member, IEEE)}{\space} received the Ph.D. degree in electronic engineering from National Chiao Tung University, Hsin-Chu, Taiwan, in 2000. He is currently a Professor in the Department of Electro-Optics Engineering, National Formosa University, Huwei, Yunlin, Taiwan, where he has been a faculty member since 2000. He is a member of the Phi-Tau-Phi Honorary Scholar Society. His research interests include optical sensors, optical signal processing and analysis with artificial intelligence approaches, and optical systems. He has been involved in multiple research projects supported by the National Science and Technology Council (NSTC), Taiwan, with contributions spanning optical measurement systems and AI-based optical analysis.
\end{IEEEbiography}
\vskip -1\baselineskip plus -1fil
\begin{IEEEbiography}[{\includegraphics[width=1in,height=1.25in, clip, keepaspectratio]{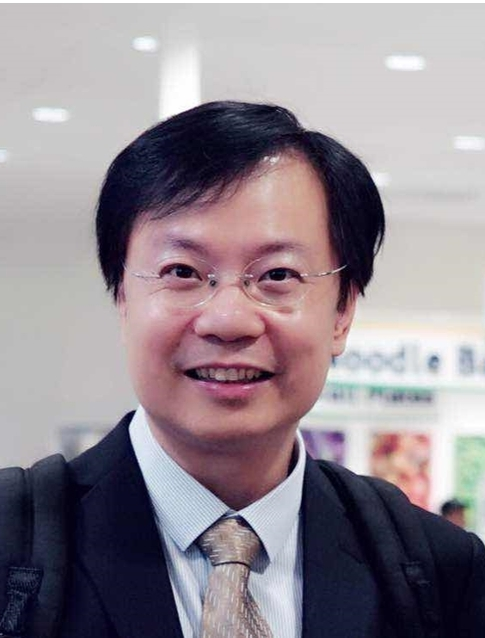}}]{Jing-Ming Guo (Fellow, IEEE)}{\space} received his Ph.D. from National Taiwan University in 2004, where he is currently a Distinguished Professor in the Department of Electrical Engineering. He was a CTO at the Industrial Technology Research Institute (2022--2023) and a Visiting Scholar at Columbia University and UC Santa Barbara. His research interests include multimedia signal processing, biometrics, computer vision, and digital halftoning. He has served as General Chair of APSIPA 2023 and Technical Program Chair of IEEE ICIP 2023, and is an Associate Editor of IEEE Trans. on Image Processing, IEEE Trans. on CSVT, and IEEE Trans. on Multimedia. He is a Fellow of AAIA and IET.
\end{IEEEbiography}
\vskip -1\baselineskip plus -1fil
\begin{IEEEbiography}[{\includegraphics[width=1in,height=1.25in, clip,keepaspectratio]{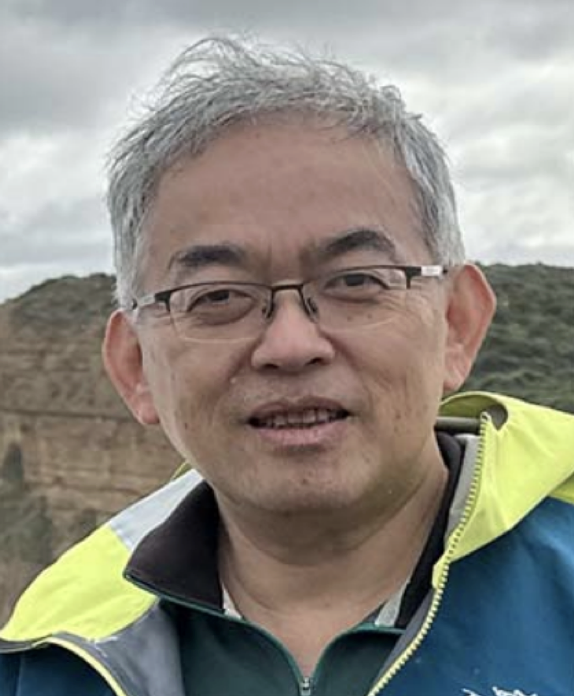}}]{Jun-Wei Hsieh (Senior Member, IEEE)}{\space} received his Ph.D. in computer engineering from National Central University, Taiwan, in 1995. He was an Assistant Professor at Yuan Ze University (1999--2009), a Visiting Researcher at the MIT AI Laboratory, and a Professor and Dean at National Taiwan Ocean University (2009--2019). He is currently a Professor with the College of AI, National Yang Ming Chiao Tung University. He has authored over 150 publications and 20 patents, with over 12,000 Google Scholar citations. His notable works include CSPNet (CVPRW 2020), PRB-FPN (IEEE TIP 2021), and SMILEtrack (AAAI 2024), with additional publications at ECCV, NeurIPS, ICIP, and ICPR. His research interests include deep learning, computer vision, object detection, multi-object tracking, neural architecture search, intelligent transportation systems, and AIoT-based smart farming. He has served as General Chair of IEEE AVSS 2025.
\end{IEEEbiography}
\end{document}